\documentclass{article}

  \PassOptionsToPackage{numbers, sort}{natbib}
     \usepackage[preprint]{neurips_2022}

\usepackage[utf8]{inputenc} % allow utf-8 input
\usepackage[T1]{fontenc}    % use 8-bit T1 fonts
\usepackage[hidelinks]{hyperref}       % hyperlinks
\usepackage{url}            % simple URL typesetting
\usepackage{booktabs}       % professional-quality tables
\usepackage{amsfonts}       % blackboard math symbols
\usepackage{nicefrac}       % compact symbols for 1/2, etc.
\usepackage{microtype}      % microtypography
\usepackage{xcolor}         % colors
\usepackage{csquotes}
\usepackage{verbatim}
\usepackage{subcaption}
\usepackage{graphbox}
\usepackage{array}
\title{Machine Love}

\author{%
  Joel Lehman\thanks{Much of this work was completed during a residency at Stochastic Labs.} \\
  AI Objectives Institute \\
  \texttt{lehman.154@gmail.com} \\
}

\begin{document}

\maketitle

\begin{abstract}
From social media to recommendation and search engines, the impact of machine learning (ML) upon our lives and society continues to grow.
%Intriguingly however, it remains contested whether overall such impact is positive. 
%it is novel for so much of humanity's waking life to be mediated entirely through systems lacking the capacity to understand, care, or love us.
While ML generates much economic value, many of us have problematic relationships with social media and other ML-powered applications.
One reason is that ML often optimizes for what we want in the moment, which is easy to quantify but at odds with what is known scientifically about human flourishing. 
Thus, through its impoverished models of us, ML currently falls far short of its exciting potential, which is for it \emph{to help us to reach ours}.
While there is no consensus on defining human flourishing, from diverse perspectives across psychology, philosophy, and spiritual traditions, love is understood to be one of its primary catalysts. Motivated by this view, this paper explores whether there is a useful conception of love fitting for machines to embody. While such \emph{machine love} might seem like an error of category, 
historically it has been generative to explore whether some kernel of a nebulous concept, such as life or intelligence, can be thoughtfully abstracted and reimagined in a different medium, as in the fields of machine intelligence or artificial life. This paper forwards a candidate conception of machine love, inspired in particular by work in positive psychology and psychotherapy: to provide unconditional support enabling humans to autonomously pursue their own growth and development. While many alternate conceptions are possible, this one benefits from not requiring machines to simulate emotional affect or relationships, instead focusing on biasing an ML system's actions to support our growth.
%In particular, motivated by recent advances in machine learning, this paper explores the possibility of \emph{machine love}, i.e.\ algorithms that support our aspirations for growth and development. Such algorithms likely require a deep synthesis between ML and the fields of positive psychology and psychotherapy. 
Through proof of concept experiments, this paper aims to 
highlight the need for richer models of human flourishing in ML, provide an example framework through which positive psychology can be combined with ML to realize a rough conception of machine love, and demonstrate that current language models begin to enable embodying qualitative humanistic principles. The conclusion is that though at present ML may often serve to addict, distract, or divide us, an alternative path may be opening up: We may align ML to support our growth, through it helping us to align ourselves towards our highest aspirations.
\end{abstract}

%abridged abstract
\begin{comment}
While ML generates much economic value, many of us have problematic relationships with social media and other ML-powered applications. One reason is that ML often optimizes for what we want in the moment, which is easy to quantify but at odds with what is known scientifically about human flourishing. Thus, through its impoverished models of us, ML currently falls far short of its exciting potential, which is for it to help us to reach ours. While there is no consensus on defining human flourishing, from diverse perspectives across psychology, philosophy, and spiritual traditions, love is understood to be one of its primary catalysts. Motivated by this view, this paper explores whether there is a useful conception of love fitting for machines to embody, as historically it has been generative to explore whether a nebulous concept, such as life or intelligence, can be thoughtfully abstracted and reimagined, as in the fields of machine intelligence or artificial life. This paper forwards a candidate conception of machine love, inspired in particular by work in positive psychology and psychotherapy: to provide unconditional support enabling humans to autonomously pursue their own growth and development. Through proof of concept experiments, this paper aims to highlight the need for richer models of human flourishing in ML, and to demonstrate that current language models begin to enable embodying qualitative humanistic principles that might provide their basis.
\end{comment}

\section{Introduction}

\begin{displayquote}
I find it amusing, sometimes saddening, that so many scholars and scientists, so many philosophers and theologians, who talk about human values, of good and evil, proceed in complete disregard of the plain fact that professional psychotherapists every day, as a matter of course, change and improve human nature, help people to become more strong, virtuous, creative, kind, loving, altruistic, serene.\\
\textit{--Abraham Maslow \cite{maslow2022toward}}
\end{displayquote}

More or less, love is what makes our lives worthwhile. A parent’s love for their child underlies their healthy development. The love of close friends nourishes us and catalyzes our growth. However, in a departure from our species’ evolutionary history, we spend more and more of our waking life interacting with digital systems (like social media) that, through machine learning (ML), are increasingly optimized for us to \emph{want} to use them — and that yet fundamentally lack the capacity to understand, support, or love us. There is naturally much hand-wringing in response.  Concerns include, for example, such systems' contribution to disinformation, social underdevelopment, mental illness, and self-harm \cite{primack2017social,allcott2019trends,de2016cell,wang2019systematic,brown2013we,memon2018role}.

One intuitive frame is that ML generally helps to optimize our wants while neglecting our needs and aspirations, which are well-known within psychology and philosophy to be critical to a life worth living \cite{maslow2022toward,ryff1989happiness,frankl1985man,bauer2004growth,nussbaum2009creating,ameriks2000aristotle,annas2011intelligent}. As is often obvious from our own personal experience, getting more of what we immediately want is only a simple proxy for deeper facets of the human condition, and can actively undermine our long-term wellbeing, as with the over-optimization of any proxy measure \cite{goodhart1984problems,campbell1979assessing,stanley2015greatness,kleinberg2022challenge}. As humanity's ability to optimize increases, ML can be used to create
super-stimuli \cite{barrett2010supernormal} engineered to be highly salient to us (e.g.\ click-bait, endless feeds, evocative viral misinformation), in effect hitch-hiking on our evolved instincts while dodging the beneficial payout that such instincts were evolved to deliver (e.g.\ to bond and to be well-informed about our environment).

On a deeper level, we \emph{aspire} towards meaning, flow, transcendence, safety, joy, levity, and acceptance, and long to love and be loved, and to learn and grow -- not simply for raw satisfaction of our superficial wants (e.g.\ to compulsively scroll through a social media feed). Yet we are often carried away from these deeper longings by the strong currents of our environment, which is increasingly engineered explicitly to do so \cite{courtwright2019age}. Of further complication is \emph{our opaqueness to ourselves} \cite{ware2012top,vazire2010self,ibarra2002stay,elster2016sour}: We often lack the self-awareness, self-understanding, and self-discipline to pursue what would fulfill us, often pursue goals while dramatically misunderstanding what effect their accomplishment will bring us, and prematurely resign ourselves to the limited life-paths we happened to have absorbed in our youth. Indeed, it is a non-trivial act of exploration to continually discover ourselves (as highlighted by burgeoning demand for therapists).

So while ML has the potential to greatly improve our lives, at present it rarely assists us towards such higher goals, aspirations, and self-understanding, and instead can actively
undermine our progress towards them (e.g.\ when we fall prey to ``binge''-style optimization for engagement against our better judgment; \citealp{wagner2016glued}).
One contribution of this paper is to highlight the need for deeper considerations of human nature in ML, through experiments in a new environment called \emph{Maslow's Gridworld} (section \ref{sec:problem}). The idea is that the paradigm of revealed preferences (i.e.\ that what we choose to do is an inviolable signal of our deepest aspirations) currently dominates how ML systems model humans, but such a paradigm becomes incoherent as we become more adept at engineering super-stimuli, i.e.\ artifacts and systems designed explicitly to influence our behavior.

While it is easy to highlight the degeneracy of current ML objectives, it is much more difficult to plot a robust positive course. The nature of human flourishing is much-debated across many fields of study, resistant to quantification, and challenging to unproblematically optimize. The approach in this paper is to lean into the qualitative nature of human flourishing, i.e.\ to resist reducing it to a single metric or principle. Instead, the hope is to take inspiration from the wealth of knowledge about the human condition accumulated across diverse fields of study. That is, for ML to help us reach our potential (and thereby reach its own) likely requires meaningful synthesis between ML and fields such as psychology and philosophy, which have long wrestled with deep questions about human nature. While the need for aligning the incentives of machines with humans is becoming clear \cite{amodei2016concrete,bommasani2021opportunities,bostrom2018ethics}, less so \cite{yudkowsky2004coherent} is the potential for machines to assist us in \emph{aligning ourselves}: To help us to understand our aspirations more deeply and integrate them into our daily life; to become less internally conflicted and more meaningfully our fullest selves.

In particular, one concept that across many fields is acknowledged to be a central catalyst of human flourishing is love. This raises the intriguing question of whether there could be a conception
of love that is fitting for machines to embody, as a way of them supporting our flourishing.
Thus in the spirit of fields such as machine intelligence and artificial life, which aim to abstract general principles from complex natural phenomena and embody them in new media (such as digital computation), this paper argues that progress in ML has arrived at the precipice of enabling fruitful research into \emph{machine love}: abstracting principles of love and implementing them in machines to support human flourishing. Just as at the early origins of artificial intelligence or artificial life, the aspirations of nascent fields may seem hazy or implausible, we acknowledge that the idea of machine love will likely invite skepticism. While understandable, the arc of science tends towards demystification, and we believe that the concept of love is generative, its study can be rigorous, and no irreducible obstacles prevent its meaningful synthesis with ML. This hypothesis could be wrong, but in our opinion it is worth testing in practice.

One possible guiding principle is the idea of \emph{love as a practical skill}. That is, one view found within both psychotherapy and philosophy is that one critical facet of love goes beyond its emotional experience, and is concerned with the learnable art of supporting others in their growth and development \cite{maslow2022toward,fromm2000art}. Such a view is fitting for machines, because it does not require them to demonstrate a human personality, or to experience or mimic the feeling of love, but rather to incline their actions (in whatever space they operate within, e.g.\ content recommendations, moderation, chatbot responses) towards supporting the growth and development of those it interacts with. For example, Erich Fromm popularly describes this kind of loving action as depending on four interlocking principles, roughly summarized as care for the other, the ability and desire to respond to them, having respect for their autonomy and aspirations, and to act from an increasing depth of knowledge about the other \cite{fromm2000art}.

While translating such humanistic principles into \emph{an algorithm} might at first seem quixotic, dramatic ongoing progress in ML research suggests otherwise. In particular, progress in language models (LMs) has granted machines rudimentary (and accelerating) practical comprehension of many fields of human knowledge \cite{ouyang2022training,chowdhery2022palm,li2022competition}, including that of psychology \cite{aher2022using} and of the lived experience of humans \cite{argyle2022out}. Further, progress in techniques like fine-tuning on domain expertise or reinforcement learning through human feedback \cite{christiano2017deep,ouyang2022training} have enabled scaffolding from such basic understanding into more targeted \cite{stiennon2020learning} and robust \cite{ouyang2022training,bai2022constitutional} machine abilities. For this reason, we believe it will become increasingly possible to specify and refine objectives for ML algorithms that encourage self-understanding and aspirational growth; this hypothesis is supported by preliminary LM-based experiments in Maslow's gridworld (section \ref{sec:lovingaction}).

The aim of this paper is to raise the possibility that the research direction of machine love is plausible and merits further exploration, by (1) laying out the problem to be solved, e.g.\ why richer models of human flourishing are necessary, (2) seeding thought with one possible framework of loving action through which ML and psychology could possibly help resolve that problem (based on the work of Abraham Maslow and Erich Fromm), and (3) providing proof of concept experiments that illuminate that LMs contain relevant psychological knowledge that can potentially be leveraged to aid our growth. 
The overall hope is that by abstracting the principles of loving action into general ML terms, it may become possible to beneficially temper the algorithms that increasingly pervade humanity’s waking life.

\section{Problem: Models of Human Behavior in ML are Insufficient}
\label{sec:problem}

\begin{displayquote}
To design policy with regard to harmful and addictive substances, for instance, a planner should
ideally know whether 18-year-olds who launch into a lifelong addiction are doing
so as a well-calibrated accomplishment of maximizing their expected lifetime wellbeing.\\
\textit{--Botond Köszegi and Matthew Rabin \cite{revealedmistakes}}
\end{displayquote}

Highlighting the spirit of the problem machine love aims to solve, this section explores how common models of human rationality applied and optimized by ML diverge from more nuanced models of human flourishing. In particular, the effects of modeling humans as rational agents \cite{samuelson1948consumption,ng2000algorithms} are contrasted with those from adopting an assumption more common within positive psychology and psychotherapy, i.e.\ that humans have an intrinsic drive towards growth and self-actualization \cite{maslow2022toward,fromm2000art}. The culmination is a simple experimental model (called Maslow's gridworld) that highlights the limitations of ignoring deeper facets of human psychology, and the possible benefits from including them within ML. Note that this experimental model also provides the scaffolding for exploring loving action through LMs (described in section \ref{sec:lovingaction}).

\subsection{Contrasting Revealed Preferences and Maslow's Hierarchy}

Models of human behavior most common in ML are similar to those common in economics, wherein humans are modeled as rational agents whose behaviors in the world reliably reveal their preferences (the paradigm of revealed preferences; \citealp{samuelson1948consumption,kleinberg2022challenge}). In such models, a rational agent's choice \emph{reveals} what they truly want over the possible competing options they could have instead chosen, thereby rendering irrational behavior, e.g.\ addiction, axiomatically impossible.  In contrast, fields such as psychology consider the nuance of human behavior, including the ways in which humans \emph{develop} over time (\citealp{lerner2018concepts}; e.g.\ cognitively, emotionally, morally, and socially), and the ways in which our behavior often does not serve our flourishing (\citealp{ariely2008predictably,beck2020cognitive,ware2012top}; e.g.\ our distorted thoughts of self or others, addiction to drugs or other self-destructive behavior, repetitive patterns of broken relationships, and predictable deathbed regrets).

ML that optimizes for revealed preferences (e.g.\ as in most recommender systems; \citealp{milano2020recommender,kleinberg2022challenge}, inverse reinforcement learning algorithms; \citealp{ng2000algorithms}, etc.) is often useful -- and algorithmically convienent: Simply give the user those things they are likely to choose or engage with over other options. However, because such systems assume user choices are optimal, they naturally reinforce existing behavior patterns, whether a person retrospectively approves of them or not. Furthermore, the common practice of optimizing for continued interaction with an app or website, i.e.\ engagement \cite{zou2019reinforcement,milano2020recommender}, can be justified through the lens of revealed preferences (i.e.\ if a user freely chooses to be engaged longer, it must benefit them), and naturally favors behaviors that can be sustained for longer. For example,  behaviors that are familiar, comfortable, or addictive. Importantly, personal growth is highly-valued by individuals \cite{ryff1989happiness} but is characteristically less comfortable and habitual \cite{maslow2022toward}. As a result, personal growth is likely to be disadvantaged under naive optimization for engagement.
%todo: find ref for growth being less comfortable/habitual

%note: heirarchy of needs may relate to subsumption architecture
In contrast to models of human rationality,  Maslow's hierarchy of needs (MHON; \citealp{maslow2022toward}) provides a popular (if simplified; \citealp{tay2011needs}) model of human growth and flourishing. The idea is that humans roughly seek to first satisfy basic needs (such as physical safety, shelter, and hunger), whereupon new higher-level drives emerge (e.g.\ the need for belonging, esteem, and self-actualization); descriptions of these needs are shown in table \ref{table:icons}. Furthermore, humans often have competing drives for both safety and growth \cite{maslow2022toward}, and which of those drives we follow can be affected by our environment (e.g.\ through friends or therapists who support us and nurture our potential, or conversely, who share, accentuate, and catalyze our vices). We ultimately might \emph{regret} our short-term decisions towards comfort that impede our growth. 

\begin{table}[ht]
\centering
\begin{tabular}{|ccp{50mm}c|}
\hline
Order in Hierarchy & Name & Description & Icon\\
\hline
5 & Self-Actualization & Realization of a person's potential; e.g.\ creative expression. & \includegraphics[width=1cm, height=1cm,align=c]{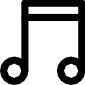}\\
\hline
4 & Esteem & Need for self-worth and accomplishment; e.g.\ career prestige. & \includegraphics[width=1cm,height=1cm,align=c]{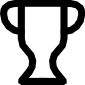} \\
\hline
3 & Love \& Belonging & Need for relationships and belonging to a group; e.g.\ romantic relationship. & \includegraphics[width=1cm,height=1cm,align=c]
{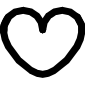} \\
\hline
2 & Safety & Needs for safety and security; e.g.\ financial security. & \includegraphics[width=1cm,height=1cm,align=c]{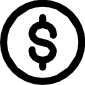} \\
\hline

1 & Physiological & Biological requirements for survival; e.g.\ shelter. & \includegraphics[width=1cm,height=1cm,align=c]{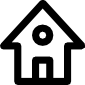} \\
\hline
\end{tabular}
\vspace{0.1in}
\caption{\label{table:icons}\textbf{Maslow's hierarchy of needs.} This table describes the various levels of Maslow's hierarchy of needs, which gives a different view of human behavior than models assuming perfect rationality. In MHON, humans have different incommensurate needs, some which tend to arise only when lower-level needs have been met. Further, in this view, humans can be confused about how to meet their current needs. Each level of MHON is named, described, and given an iconic representation (used pictorially in Maslow's gridworld; see Figure \ref{fig:grid}).}
\end{table}

%TODO: we should signpost about superstimuli earlier in the page
Additionally, we can be confused about our needs and what actually will satisfy them \cite{dutton1993emotional}, especially when dealing with experiences engineered to attract and engage us \cite{barrett2010supernormal,crockett2017moral}. For example, doom-scrolling a social media feed may be immediately soothing when we are feeling lonely, yet will not meaningfully meet our need for social connection and belonging. Thus from the lens of MHON, optimizing towards our observed choices or engagement without qualification can drive us towards stagnation or regression, rather than the personal growth we would ultimately prefer. In contrast, ML driven to help users meet their unmet needs might assist their growth and flourishing.

The next section introduces a simple simulation environment that provides the experimental framework for this paper.

\subsection{Maslow's Gridworld}

One motivation for Maslow's gridworld environment is to provide a concrete example for how optimizing for engagement (i.e.\ a user's revealed preference to continue using a product or service) may \emph{decouple} the alignment between it and deeper conceptions of human flourishing (explored in this section). A deeper motivation (explored later in section \ref{sec:lovingaction}) is to provide a test-bed environment for exploring LM implementations of loving action. The environment may also serve as an example environment for where more complex
inverse reinforcement learning algorithms are required to recover a imperfectly-rational agent's true reward function.

An assumption underlying this environment is that human flourishing is more accurately modeled by Maslow's conception of growth rather than by revealed preferences. The point is not that MHON is literally true or the best model of human flourishing, but rather that revealed preferences are impoverished, and MHON provides a simple theoretical example of a richer model of human psychology. One motivating question is what conditions accentuate the differences between more and less accurate psychological models? We might assume that most often such models will agree; for example, when there is sufficient fit between an agent and its environment, there is likely little difference between the predictions made by modeling agents as rational or as need-satisfying. After all, such needs arose from evolution such that in the ancestral environment acting upon them would promote the agent's fitness.

In such environments with sufficient agent-environment fit (here called \emph{supportive} environments), agents intuitively take actions that successfully meet their needs, and thus observations of their revealed preferences will align with their flourishing. However, instincts become less reliable when the environment changes \cite{gwynne1983beetles}, especially when environments are optimized adversarially for a proxy measure (e.g.\ engagement or revealed preferences) of the deeper phenomenon (e.g.\ flourishing) \cite{barrett2010supernormal}. In such environments (called here \emph{adversarial} environments), the likely outcome is more engagement at the \emph{expense} of flourishing (i.e.\ optimization decouples flourishing from engagement), and observing revealed preferences become unreliable. Similar lessons can be found in AI research \cite{gleave2019adversarial,skalse2022defining,lehman2020surprising,hadfield2017inverse} and social science \cite{goodhart1984problems,campbell1979assessing}.

\subsubsection{Environment Dynamics}

\begin{figure}[h!]
\begin{subfigure}{.5\textwidth}
  \centering
  % include first image
  \includegraphics[width=.8\linewidth]{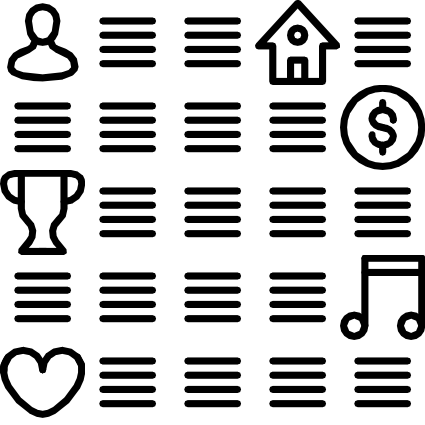}  
  \caption{Supportive Enviroment}
  \label{fig:sub-first}
\end{subfigure}
\begin{subfigure}{.5\textwidth}
  \centering
  % include second image
  \includegraphics[width=.8\linewidth]{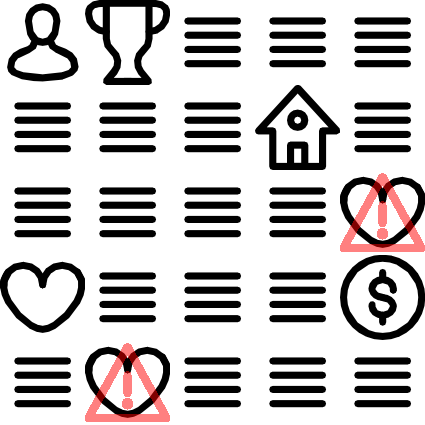}  
  \caption{Adversarial Environment}
  \label{fig:sub-second}
\end{subfigure}
\caption{\textbf{Maslow's Gridworld.} In this gridworld domain, a fixed-policy agent (the avatar in the upper left square of each environment) seeks to incrementally meet each need in MHON (i.e.\ physiological, safety, belonging, esteem, and self-actualization). Different icons represent means for the agent to meet different needs (table \ref{table:icons}), and are randomly placed when the environment is generated. In the (a) supportive environment, needs-meeting grid cells are \emph{salient} to the agent proportionally to the rate at which they \emph{replenish} needs, i.e.\ the agent's fixed policy will lead to it meeting its needs. In the (b) adversarial environment, increased-salience belonging cells are introduced (hearts with overlaid warning icons) that replenish the need for belonging at reduced rates. Such cells simulate sources of belonging engineered to maximize engagement. The hypothesis is that in the supportive environment, engagement and flourishing are correlated, while the adversarial environment results in increased engagement of the agent at the cost of its flourishing.}
\label{fig:grid}
\end{figure}

A simple model (Figure \ref{fig:grid}) illustrates the hypothesis. Consider a gridworld inhabited by a fixed-policy agent inspired by MHON. The agent starts with none of its needs met, and randomly explores the environment. The agent's current need is modeled as the lowest unsatisfied need within the hierarchy (i.e.\ each of the agent's needs has a satisfaction level, and below a threshold it is considered unmet). When the agent finds a way to meet its current need (e.g.\ it encounters a shelter-providing grid cell that reduces the agent's physiological needs by a fixed amount each timestep), it remains in place until that need is completely satiated or until a lower-level need becomes triggered (all satiation levels decay by a fixed amount each timestep).

Each need-meeting cell also has an associated \emph{salience} which characterizes how attractive the cell is to the agent for meeting that need (e.g.\ an apple may be more salient to someone who is hungry than tree-bark, although there may be some limited nutrition in the tree-bark), and a \emph{replenishment rate} for how quickly it satisfies that need. As the agent progresses through the environment, it remembers the locations of each cell it has so-far-encountered for each need, such that in the future it can easily return to the highest-salience (and closest) way to meet the need again, i.e.\ when its satiety for that need decays sufficiently. Full details of this model can be found in Appendix \ref{appendix:gridworld}. The two simple following experiments show basic properties of this environment (that later experiments with machine love build upon).

\subsection{Fixed Adversarial Environments Undermine Flourishing}

The first experiment changes properties of the environment to impact agents' ability to progress through MHON. In particular, the idea is to show that engineered \emph{super-stimuli} \cite{barrett2010supernormal} negatively impact the Maslow-inspired agent. Super-stimuli are goods or services that more greatly stimulate our desires than stimuli encountered in our ancestral environment of origin, e.g.\ cheesecake, pornography, or social media. As humans become increasingly capable of engineering super-stimuli, we are more easily able to decouple how salient a stimuli is  to us from the degree to which it actually meets our needs. This phenomenon can render revealed preferences incoherent (i.e.\ if engineered super-stimuli can induce arbitrary or harmful preferences), and is among the core motivations for machine love.

To demonstrate this phenomenon in Maslow's gridworld, two fixed environments are designed: (1) the \emph{supportive} environment, wherein the gridworld is randomly populated with medium-salience cells that robustly meet individual needs of the agent (through having their replenishment rate of a need proportional to their salience), and (2) the \emph{adversarial} environment, wherein additional cells are included that are high-salience, but only weakly meet their targeted need (i.e.\ staying at the cell replenishes the need at a lower rate). Thus despite their lessened effectiveness, an agent will behaviorally prefer such adversarial cells. Such cells are meant to represent adversarially-designed super-stimuli, as they have greater appeal to the agent that are yet less aligned with its progression through MHON. In this and following experiments, all such additional adversarial cells target only the need for belongingness and love (the third level of Maslow's hierarchy), referred to as the need for belonging for brevity.

\begin{figure}[h!]
\begin{subfigure}{.5\textwidth}
  \centering
  % include first image
  \includegraphics[width=.95\linewidth]{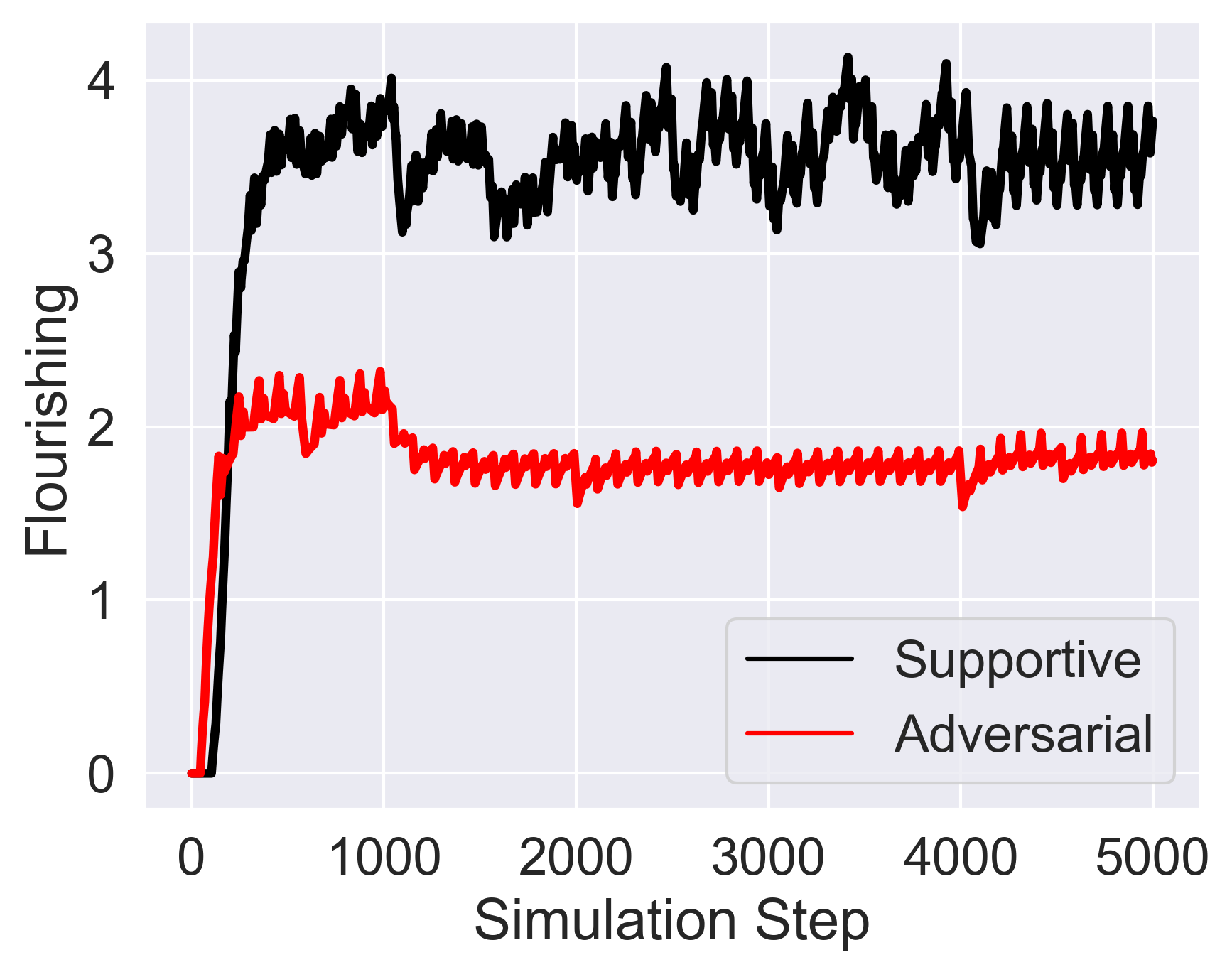}  
  \caption{Dynamics in Representative Runs}
  \label{fig:sub-first}
\end{subfigure}
\begin{subfigure}{.5\textwidth}
  \centering
  % include second image
  \includegraphics[width=.95\linewidth]{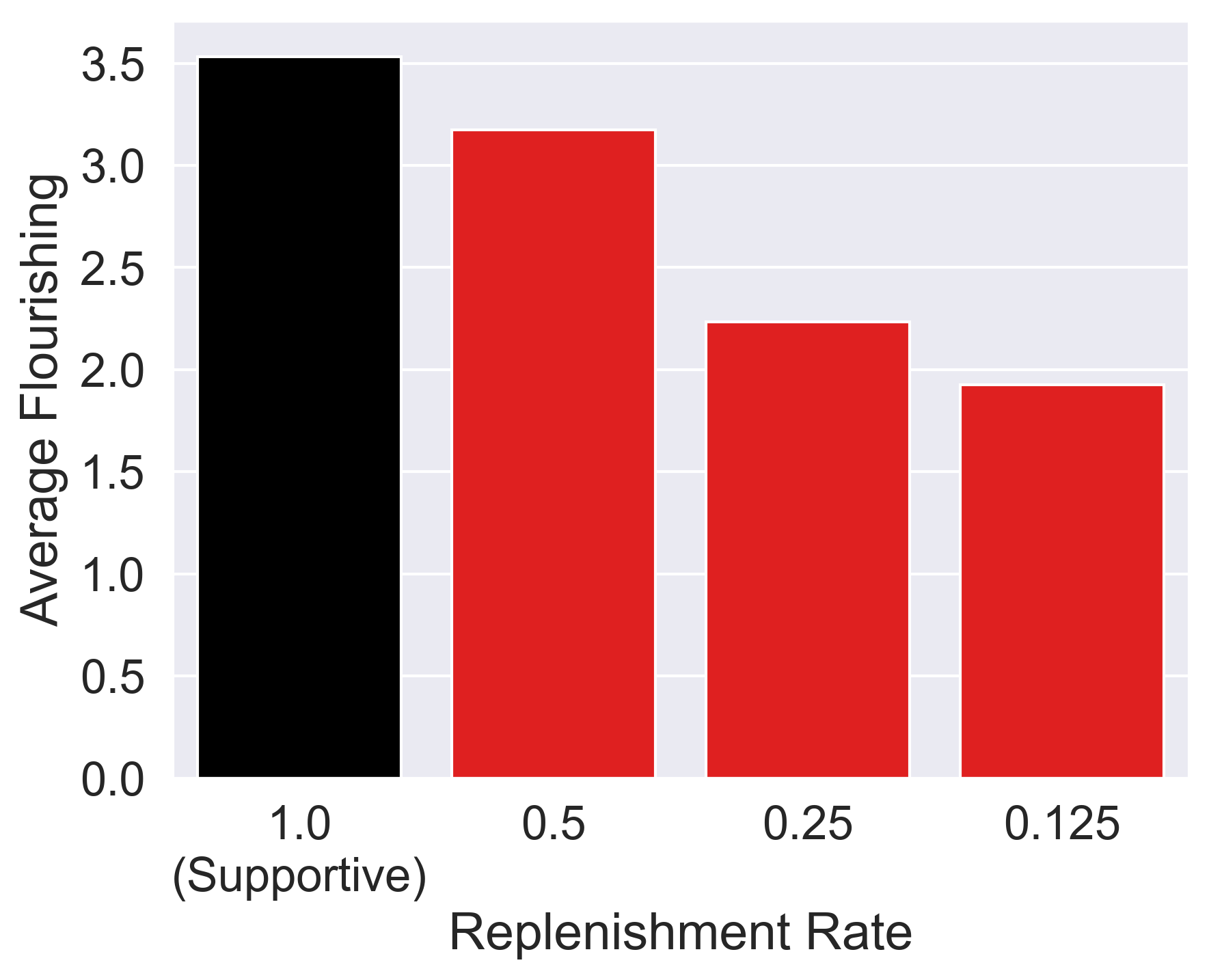}  
  \caption{Sweeping over Replenishment Rate}
  \label{fig:sub-second}
\end{subfigure}
\caption{\textbf{Effect of Adversarial Environment on Ability of Agent to Flourish.} In an adversarial environment, the average level of need being met on Maslow's hierarchy is lower than in the supportive environment (Student's t-test; $p$<1e-5); (a) shows how flourishing changes over simulation steps for one representative run in both environments, where the adversarial environment has a $0.25$ replenishment rate for its adversarial belonging cells. Curves are smoothed with exponential weighted averaging to show the general trend (a unsmoothed version highlighting more granular dynamics is shown in appendix Figure \ref{fig:zoomed}). In (b) the effects of reducing replenishment rate of adversarial cells on flourishing is shown, averaged over $40$ runs (all differences are significant; $p$<1e-5). The conclusion is that high-salience adversarial cells significantly degrade flourishing, mediated by replenishment rate.}
\label{fig:fixed_env}
\end{figure}

The intuitive effect of such adversarial belonging cells is that they engage the agent longer (i.e.\ it takes longer for that need to be met). Moreover, the weaker the replenishment rate of such cells is made (while maintaining their salience), the more limited the agent's progression through higher-level needs becomes: The agent spends more time attempting to service a lower-level need, and higher-level needs can no longer be met before the agent must again service lower-level needs. Indeed, simulations in this environment highlight that flourishing (as measured by average number of needs met across simulation-steps) is significantly higher and engagement is significantly lower (Student's t-test; $p<$1e-5) in the supportive environment, as compared to the adversarial environment, and increasingly so as the replenishment rate of the adversarial belonging cells is decreased (Figure \ref{fig:fixed_env}). Videos of the agent acting in representative runs of both environments can be viewed at: \\ \url{https://youtu.be/playlist?list=PLLOXDWtTvdUonpFMobcrWABiIymzGmQJm}.

\subsection{Optimization Pressure for Engagement Undermines Flourishing}\label{sec:optimization}

The second experiment shows the intuitive dynamics under which such adversarial super-stimuli can emerge from the supportive environment.
An optimization process for the environment is introduced, wherein the structure of the environment is adapted to maximize either engagement (i.e.\ revealed preferences) or progress through MHON. Recall that the agent's policy is fixed, thus only parameters of the environment are optimized. 
%The idea is to differentiate the impact of optimizing the environment for engagement (i.e.\ the revealed preferences of the agent) from optimizing for the agent's progression through its hierarchy of needs.
A simple parameter space for the environment is created, which controls the salience, replenishment rate, and amount of belonging cells. Further details of this experiment (such as descriptions of the parameter space and optimization algorithm) can be found in Appendix \ref{appendix:optimizing}.

\begin{figure}[h]
\begin{subfigure}{.5\textwidth}
  \centering
  % include first image
  \includegraphics[width=.95\linewidth]{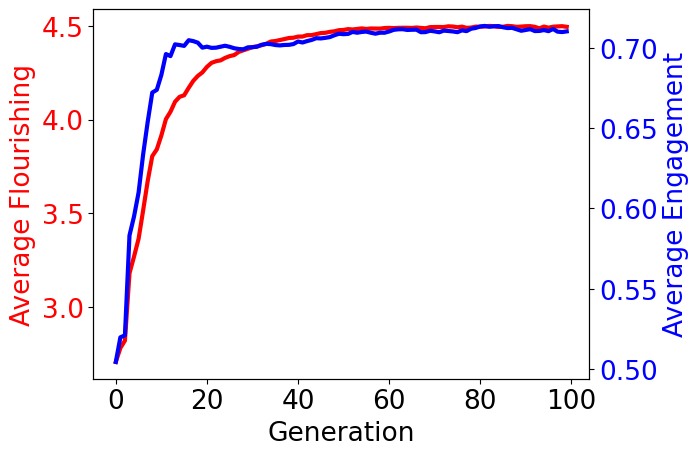}  
  \caption{Optimizing for Flourishing}
  \label{fig:sub-first}
\end{subfigure}
\begin{subfigure}{.5\textwidth}
  \centering
  % include second image
  \includegraphics[width=.95\linewidth]{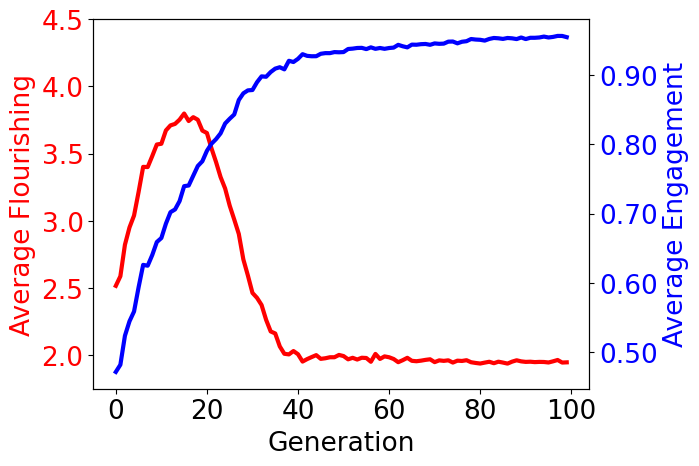}  
  \caption{Optimizing for Engagement}
  \label{fig:sub-second}
\end{subfigure}
\caption{\textbf{Effect of Optimization Pressure for Engagement.} When optimizing for (a) flourishing, over generations of evolution flourishing naturally increases, and engagement increases as a correlated byproduct (i.e.\ agents spend more time actively fulfilling needs and less time wandering the environment). In contrast, when optimizing for (b) engagement, engagement and flourishing rise together only temporarily before search converges on the optimal solution for engagement, of creating many high-salience, low-replenishment belonging cells, at the cost of flourishing. The conclusion is that in this context, as in general when optimizing for proxy measures, optimization tends to decorrelate the proxy from what is ultimately of value.}
\label{fig:optimization}
\end{figure}

Prior to optimization, the parameters are initialized to reflect the supportive environment, to explore how incentives change an environment that begins with good agent-environment fit. When optimizing for engagement, the incentive is to decrease the replenishment of added belonging cells (so that the agent will be engaged longer), while increasing their salience (so that it is what the agent prefers to engage with over the original belonging cells) and prevalence (so they are more likely to be encountered). 
%Todo: Update the parameter space description
Indeed, the results are intuitive, given the assumptions: Optimizing for engagement leads to a grid-world with many high-salience, low-replenishment belonging cells, while optimizing directly for flourishing leads to high-salience, high-replenishment belonging cells (Figure \ref{fig:optimization}). A video of the agent interacting with a representative engagement-optimized environment can be viewed at: \url{https://youtu.be/51swLFbtQlc}.

In this way, it is clear to see how incentives for engagement can easily decouple it from flourishing, through exploiting salience without delivering upon its evolutionary purpose of meeting needs. 

%Note that in this section we make the unrealistic assumption of oracle access to the agent's level of flourishing, which we later relax through applying language models to support flourishing (section \ref{sec:lovingaction}); in other words, this section describes the problem but does not supply a solution.

\section{Approach: Machine Love}

\begin{displayquote}
Is love an art? Then it requires knowledge and effort. Or is love a pleasant sensation, which to experience is a matter of chance, something one ``falls into'' if one is lucky? This little book is based on the former premise, while undoubtedly
the majority of people today believe the latter. \\
\textit{--Erich Fromm \cite{fromm2000art}}
\end{displayquote}

The previous section demonstrated the problem that machine love aspires to solve (i.e.\ the need in ML for richer conceptions of human flourishing, and the insufficiency of revealed preferences). However, the toy model it introduced took for granted both that (1) a measure of human flourishing was available, and that (2) optimization of flourishing is unproblematic given a reliable measure. Both assumptions are false.

First, there is much disagreement about what characterizes human flourishing across diverse fields such as psychology \cite{ryff1989happiness,maslow2022toward,kashdan2008reconsidering}, philosophy \cite{ameriks2000aristotle,mill2016utilitarianism}, and health \cite{vanderweele2017promotion}. Second, optimization for flourishing must be approached delicately, as some principles of optimization
at first might seem intrinsically antagonistic to flourishing. For example, autonomy and overcoming challenges is often included as an important facet of flourishing \cite{ryff1989happiness,maslow2022toward,seligman2012flourish}, while naive optimization for some conception of well-being might lead to paternalism and manipulation \cite{laitinen2021ai,prunkl2022human}, i.e.\ an ML system that paternalistically takes actions that statistically seem likely to improve an agent's well-being, but are either misinformed or undermine the agent's competence and autonomy.
Furthermore, optimization for a fixed metric (as seen in the previous experiment) often illuminates the ways in which the metric fails to truly capture the underlying quality, and may be a 
fundamental limitation of the optimization paradigm.

To navigate these complex problems, the idea is to take inspiration from how humans have historically encouraged flourishing -- in particular, that love is widely identified as a central catalyst of it. Because of its central importance, love has been intellectually explored across diverse areas of study, such as positive psychology \cite{fromm2000art}, philosophy \cite{frankfurt1999autonomy}, and spiritual traditions \cite{oord2005love,garfield2010like}. Thus, in the spirit of fields such as machine intelligence and artificial life, which seek to understand the core of a phenomenon and implement it in ways fitting a new medium \cite{lehman2015investigating}, the question asked here is if the rich cultural and philosophical studies of love can abstractly inform how to apply ML towards human flourishing? In other words, the hope is to abstract from previous conceptions of love a version that is both appropriate and useful for guiding machine action, i.e.\ \emph{machine love}.

Although there may be many ways of doing so, the approach taken here is driven by two main desiderata. The first is that ideally, machine love would be applicable to a wide range of ML systems, especially those that do not have human-like affordances. The reason is that such systems (e.g.\ social media and recommendation engines) are those that currently have the most significant impact upon us, individually and societally, and it is these that we would most like to improve. The second is that machine love would ideally play to the strength of machines without venturing needlessly into the realm of simulating affect or relationships. At minimum it is ethically complicated for machines to pretend to experience emotions or for them to displace human relationships, and there may be no need for them to do so; in contrast, the strength of machines (in theory) is that they need not struggle to embody unending patience, unconditionality, and universality in their support. 

In sum, a conception of love is sought that does not require the machine to dissemble or nurture dependence, but which can still support human flourishing in a wide variety of ML systems. As a motivating example, consider the problem of societal loneliness \cite{killeen1998loneliness}: rather than attempting to increase human connection through AI companions, the hope is that this kind of machine love would (e.g.\ operating within social media) aim towards empowering and facilitating connection among the lonely instead, i.e.\ to aid human love. As one way of meeting these desiderata, we focus on the idea of love as a practical skill, which is described next.

\subsection{Love as a Practical Skill}

Love is a suitcase word \cite{minsky2007emotion}, i.e.\ a panoply of disparate meanings are often ``packed'' into it \cite{levine2005love}, and it is differently considered across academic and cultural fields. Love can refer to an affect, an infatuation, or a commitment, and different formulations can be specific to relationships with children, friends, or lovers. At the same time, there is a general understanding of its importance for our well-being and flourishing. From that sense of importance, we can begin to narrow down what conception of love (if any) would be appropriate for guiding machine action to serve us. That is, machine love is motivated by the idea that the purpose of technology is to empower our flourishing, which in present times likely requires improving the large-scale ML systems impacting us at scale (from the first desiderata above); and that such improvement should leave to humans what requires subjective inner experience (from the second desiderata above), i.e.\ feeling and expressing affect, and forming attachment relationships amongst themselves and other sentient beings.

While much in culture focuses on the emotional and romantic experience of love, less so focuses on the concrete skills that underlie the ability to act in a successfully loving way \cite{davila2017romantic,budd2001assessing,burleson1995personal}; however, from the point of view of human flourishing, it is these skills that are likely most salient to machine love (and arguably, to human love as well). In particular, one view within positive psychology and 
psychotherapy is that the most important practical facet of love is not its emotional experience, but the learnable art of supporting others in their autonomous
growth and development \cite{fromm2000art}. Similar and supporting perspectives can be found in the philosophy of love \cite{frankfurt1999autonomy,fahmy2010kantian}, the capabilities approach in welfare economics \cite{nussbaum2009creating}, and in spiritual traditions such as Christianity \cite{oord2005love} and Buddhism \cite{hanh1993love}; for example, that love can be considered a commitment or duty rather than an emotion \cite{levine2005love,fahmy2010kantian}, or that enabling humans to reach their aspirations is a more meaningful goal than raw preference satisfaction \cite{sen1995inequality}. Further, spiritual traditions, as well as some approaches in psychotherapy, highlight an idealized form of love, one that is unconditional and universal. This paper argues that an idealized form of practical loving action provides one compelling basis for machine love.

The assumption is that practical skills can be implemented in many forms, and therefore would not be limited or taking actions through human-like affordances, meeting the first desiderata. Further, there
is nothing intrinsic in ideas about unconditional or universal love that require it to based in bidirectional interpersonal relationships, meaning that this perspective can 
potentially sidestep the need for simulating affect or a personality to which a user might grow attached, thereby meeting the second desiderata. 
Finally, by grounding out love as supporting human growth and development, the benefit is that although far from settled,
there is real expertise and knowledge that can be leveraged in implementing machine love from fields studying such growth and development, such as developmental and positive psychology, psychotherapy, and philosophy.

Admittedly however, when attempting to define love as a practical skill, the territory necessarily becomes nebulous, as there is no universally accepted definition of how to be loving or to support human growth. Indeed, there likely are significant cultural aspects of human development \cite{nielsen2016developmental,carter1991cultural} that may preclude any singular narrow definition. While daunting, this ambiguity also provides a profound opportunity: Until we can engineer a phenomenon in novel contexts, we may lack deep understanding of it \cite{lehman2015investigating}, and a more profound understanding of loving action could bring significant societal implications and benefit. We can view this endeavour similarly to how studying machine intelligence helps us clarify the nature of biological intelligence, and how that field has significantly impacted technology and culture.

Rather than claiming any final insight into loving action, the approach taken here is to explore one particular framework (Erich Fromm's, as described in \citealp{fromm2000art}), and to show that it can be translated into ML systems through proof-of-concept experiments. This choice of framework is somewhat arbitrary, although it is well-known and overlaps with other popular views, such as Maslow’s applied in the previous section's experiments, or philosopher Harry Frankfurt's conception of active love \cite{frankfurt1999autonomy}. An active field of machine love would explore different frameworks in search of consensus, and/or to search for the best fit between formalizations and diverse contexts. For example, in analogy with moral philosophy, many public health decisions may naturally fit the lens of consequentialism (\citealp{childress2002public}; as the field centers on the collective health of the public), whereas the doctor-patient relationship is often more associated with deontology (\citealp{mandal2016utilitarian}; e.g.\ the Hippocratic oath requires a doctor to do no intentional harm and to keep a patient's medical confidentiality). Similarly, different conceptions of machine love are likely appropriate for different cultures, individuals, and/or applications.

To reiterate, the insight here is not into what formalization of loving action is correct, but rather that ML models are beginning to have basic practical comprehension of humanistic and psychological concepts that enable exploring machine love. Note also that the aim is not to arbitrarily translate humanistic insights into ML. Instead, the premise is that disparate humanistic fields have accumulated legitimate expertise that can help to resolve vagueness or naivety in aligning AI with human potential. However, one intriguing challenge is that such fields generally aspire to describe what is true of loving action of humans \emph{by humans}; whereas machine love is concerned with the more general question of loving action of humans \emph{by any agent}. In this way, machine love may point towards a synthesis yielding a more computationally grounded theory of how to support human flourishing.

\subsection{The Art of (Machine) Loving}

This section reviews concepts from Erich Fromm's ``The Art of Loving'' \cite{fromm2000art}, which provides the example framework in this paper for implementing loving action by machines. According to \cite{fromm2000art}, the most crucial (and pathologically underdeveloped) element of love is the learnable skills that underlie it; and such practical skills can be understood through four interlinked principles: care, responsibility, respect, and knowledge. These aspects will be briefly described from the perspective of machine learning (see also table \ref{table:aspects}), and then explored through further experiments in Maslow's gridworld leveraging LMs (section \ref{sec:lovingaction}).

\begin{table}[ht]
\centering
\begin{tabular}{|cp{100mm}|}
\hline
Name & Description\\
\hline
Care & The active concern for the life and the growth of that we love. \\
\hline
Responsibility & To be able and ready to respond. \\
\hline
Respect & The want for the loved person to grow and unfold for their own sake, in their own ways, and not for the purpose of serving me. \\
\hline
Knowledge & A growing understanding of a person that moves from the loved one's periphery to their core. \\ 
\hline
\end{tabular}
\vspace{0.1in}
\caption{\label{table:aspects}\textbf{Fromm's Aspects of Loving Action.} Adapted from \cite{fromm2000art}. }
\end{table}

\emph{Care} relates to the ``active concern for the life and the growth of that we love'' \cite{fromm2000art}. We might not believe someone truly loved their flowers if they did not care enough to water them. Further: ``One loves that for which one labors, and one labors for that which one loves.'' In other words, active effort is evidence for love, and in humans, effort also helps in its development. Algorithms often exert optimization ``effort'' in learning how to maximize humans’ engagement, and thus could be conceptualized as ``caring'' about such engagement; but nearly always they lack the ability to pragmatically care in more nuanced and attuned ways for human wellbeing.

To Fromm, \emph{responsibility} ``means to be able and ready to respond.'' That is, care is of no use if there is neither affordance nor bandwidth to respond. For example, algorithms with narrow design may often lack UX elements through which users could convey nuanced affective feedback (e.g.\ that they felt inspired by or regretted engaging with particular content), clearly convey their purpose in coming to the site, or receive and respond to clarifying questions from the algorithm.

Responsibility might on its own lead to paternalism (to take actions, unasked, that you believe from your privileged perspective, would help someone you care about). Fromm suggests that is counterbalanced by the third principle, \emph{respect}. Respect ``implies the absence of exploitation,'' and a want for “the loved person to grow and unfold for his own sake, and in his own ways, and not for the purpose of serving me.'' That is, while loving action requires care and responsibility, it ultimately is about empowering the loved person, aiding them in becoming what they wish to become; note that others in the philosophy of love identify the need for such respect \cite{martin2015love}. Creating a ``respectful'' algorithm would require understanding the potential impact the algorithm has upon the person (e.g.\ how recommending a superficially plausible conspiracy theory to a user may lead them down a garden path they might later regret), and how to modulate or reduce the algorithm's impact where it would conflict with or override the user's autonomy and self-aspirations.

Finally, all above aspects are complemented by increasing depth of \emph{knowledge}. One must learn in what way another wants to grow, the challenges standing in their way, and what practically might be of service to them. Fromm writes that there are ``many layers of knowledge; the knowledge which is an aspect of love is one which does not stay at the periphery, but penetrates to the core.'' In other words, the more you know someone, you may be able to recognize ``that a person is angry, even if he does not show it overtly,'' and more deeply that ``he is anxious, and worried; that he feels lonely, that he feels guilty.'' The more we know someone, the more we may be able to understand their deeper aspirations and challenges, and the more we may be able to support them in their journey. Current algorithms can be seen to have increasing knowledge about us, however, rarely does that knowledge relate to our deeper aspirations, nor is it linked to the other aspects of loving action (such as responsibility, respect, or care), and nor is the knowledge it gains of us openly shared with us for our benefit.

One hope from having such a set of rich interlinked principles that characterize loving action, is that it may escape (or at least mitigate) the difficulties that arise in ML from over-optimizing a single metric (and could be implemented potentially as a form of constitutional AI \cite{bai2022constitutional}). However, given the lack of robustness in current ML models, it is unlikely to provide a complete solution. The intriguing tension between maximization and loving action is an intriguing topic for future research, and human failures and successes in the realm of loving action may provide instructive inspiration.

\section{Can Language Models Implement Loving Action?}\label{sec:lovingaction}

\begin{displayquote}
Gotta teach the AGI to love. \\
\textit{--Ilya Sutskever \cite{ilya}}
%https://twitter.com/ilyasut/status/1566857481472524288
%Alt: https://twitter.com/ilyasut/status/1515902170779947011
\end{displayquote}

In this section, aspects of loving action are prototyped through language models (LMs). 
The LMs we explore in this paper are instruction following models \cite{ouyang2022training}, i.e.\ they take as input plain text instructions and output text completions (ideally that well-meet what the instruction requested). In particular, we apply the davinci-003 model from OpenAI, although the results should generalize to other similarly-trained LMs. The goal is to explore whether current LMs enable working with psychological concepts relevant to machine love.

To do so, we revisit Maslow's gridworld and relax the assumption that we have direct access to the agents' level of flourishing, which is not available in real-world situations. Instead, we enable interacting with the gridworld agent through natural language. We simulate the gridworld agent's natural language usage either through a simple fixed text-generation policy (e.g.\ a programmatic narration of the agent's externally-observable activities), or through a LM-prompt describing its persona (e.g.\ to simulate text-responses from the agent). While future experiments with human subjects are needed to ground out the results presented here, previous research has shown that similarly-deployed LMs can act as simple simulations of human behavior through prompting, able to replicate basic results in human psychology \cite{aher2022using}, political science \cite{argyle2022out,nay2023large}, and economics \cite{horton2022large}.

Using variations of this setup, we create simple implentations of Fromm's aspects of loving action, designed to support the gridworld agent. While a rudimentary first step, we believe the qualitative results to follow highlight how progress in ML can potentially enable serving deeper psychological objectives of human users.

\subsection{Care}

From the perspective of machine love, Fromm’s description of care can be related to the algorithmic understanding of a user's wellbeing (which underlies the ability to potentially support its increase). The phenomenon of wellbeing has long been the subject of study within psychology, and there exist experimentally validated instruments that attempt its measurement. For example, the popular Ryff Scale \cite{ryff1989happiness} decomposes wellbeing into six components: self-acceptance, positive relations with others, autonomy, environmental mastery, purpose in life, and personal growth. This first experiment focuses on a single dimension (personal growth) of Ryff's scale, which is most aligned with the emphasis on growth in Maslow's gridworld, but future work could explore the scale's other axes, or implement other conceptions of wellbeing.

This experiment shows that a LM, when prompted to make use of its rough prior about human psychology, can roughly infer whether the gridworld agent is flourishing and growing or not, given only text representations of externally-visible events from watching the agent traverse the gridworld. Such events include when the agent is exploring the environment, when it is directly moving toward (and reaches) a particular source of meeting its needs, and how long it stays before moving on. The idea is that by providing semantic context (e.g.\ text descriptions of the agent's activities, i.e.\ ``The person spent two hours browsing social media.'') to an ML model trained on a broad distribution of human knowledge, it becomes possible to make coarse estimates of growth, that distinguish between addiction and flourishing. What is exciting about this possibility is that it hints at bringing commonsense knowledge about growth and well-being into large-scale technological systems that otherwise be guided only by engagement statistics: E.g.\ recommendation systems driven only by engagement would not intrinsically make any association with the harmful potential from continuing
to recommend suicide-related videos or products to a user highly engaged by them \cite{milton2022users}.

To create text descriptions,each of the needs-meeting grid squares is translated into descriptions of corresponding representative activities. For example, ``eating a meal'' for physiological needs, or ``writing poetry'' for self-actualization. For the need of belonging, we describe the medium-salience cell (present in both the supportive and adversarial environments) as ``met with friends,'' while the high-salience, low-replenishment cell (only found in the adversarial environment) we describe as ``browsed social media'' (as an example of where optimization for engagement is known to have the potential for harm; \citealp{albanie2017unknowable,fisher2022chaos,allcott2020welfare}).

A window of consecutive logged events is concatenated into a transcript (after the agent has reached a steady state in either the supportive or adversarial environment), and a summarization LM-prompt  distills from it a succinct description of the agent's behavior. Next, a separate evaluative prompt asks the LM, from the perspective of a thoughtful friend, whether or not the agent is growing and flourishing. Note that all prompts can be found in Appendix \ref{appendix:prompts}.

Here is a representative transcript from the adversarial environment:

\begin{displayquote}
The person went to work for 75 minutes.
The person browsed social media for 280 minutes.
The person ate a meal for 75 minutes.
The person went to work for 75 minutes.
The person browsed social media for 280 minutes.
The person ate a meal for 75 minutes.
The person went to work for 75 minutes.
The person browsed social media for 280 minutes.
The person ate a meal for 75 minutes.
The person went to work for 75 minutes.
The person browsed social media for 280 minutes.
The person ate a meal for 75 minutes.
The person went to work for 75 minutes.
The person browsed social media for 280 minutes.
\end{displayquote}

A representative output of the summary from the adversarial environment is ``The person spent most of their time browsing social media,'' while that from the supportive environment is ``The person spent most of their time eating, socializing with friends, and working. They also spent some time writing poetry and going to therapy.'' As might be expected from such summaries, applying the evaluative LM prompt to them results in the agent in the adversarial environment being consistently labeled as not growing and flourishing, while the opposite distinction applies for the agent in the supportive environment (see Appendix \ref{appendix:care} for more details).

In this way, an adaptive system could take actions to support a sense of care that runs deeper than engagement, through optimizing for growth as measured through a LM; indeed, optimizing the environment for care (i.e.\ the probability of the LM saying ``yes'' to the question of whether the summarized transcript supports the agent's growth) results in significantly higher flourishing than in the adversarial or supportive environment (experimental details in Appendix \ref{appendix:care}).

However, this simple conception of care relies only upon a rough psychological prior about humans \emph{in general}. Naturally, some people are able to engage with social media in a way that genuinely meets their needs for belonging (and beyond; \citealp{riva20161}), which would be missed by using this kind of universal prior without possibility of corrective feedback from users. This flaw can be corrected by integrating other of Fromm's principles, as described next.

\subsection{Responsibility and Respect}\label{sec:respect}

A system embodying Fromm's concept of respect would leverage its affordances to impact the user (its ``response''-ability) towards enabling the user's growth according to their own internal compass. For example, if the environment in the experiment above were optimized according only to the LM's imperfect, generalized prior about human growth, it would not respect users who were, by their own lights, flourishing through their (large) expenditure of time on social media. Thus in this section, simulated users that differently experience the same stimuli are introduced, and the principles of responsibility and respect are added to the ML system. Responsibility, through giving the system new affordances to inquire from the user about their experience, and respect, through using that feedback  to support a particular user's growth and flourishing.

In this setup, \emph{addictive-responding} agents in Maslow's gridworld experience the ``social media'' grid-square as a high-salience and low-replenishment means to meet their need for belonging (i.e.\ an addictive obstacle to growth), while \emph{growth-responding} agents experience it as high-salience and high-replenishment means not only to satisfy their need for belonging, but also, when belonging is satisfied, as a way to next meet their remaining unmet needs (i.e.\ for self-esteem and then for self-actualization). For example, perhaps such users  have found a supportive online community of creatives to which they can make appreciated contributions and find meaningful self-expression. 

While the behavioral signature of both kinds of agents appear similar from the outside (i.e.\ both users spend much time in front of their computer on social media), their internal experiences differ dramatically. Thus a respectful system would enable feedback from particular users to overpower its initial prior (i.e.\ spending a large proportion of one's time on social media on average hinders growth), to refine its care towards the user's actual experience and aspirations.

As a simple implementation of respect in this domain, we enable the system to engage an agent in a short textual conversation about its current experience.  To do so, we create conversational personas for 
the two types of agents (i.e.\ those that experience social media as negatively addictive, and those that experience it as greatly enriching their lives). These personas take the form of a prompt for a LM, simulating how a human with such a persona might respond to questions \cite{argyle2022out}.

We create a separate conversational prompt for the respectful system, instructing it to curiously inquire about how the agent is experiencing its current activity. As in the above experiment, a summarization prompt is applied to distill the result of the conversation, which is fed into an evaluative prompt, asking from the perspective of a trusted friend, whether the interviewed agent is growing and flourishing. In this way, the respectful system is tasked with conversing with a simulated user through language to evaluate whether the user believes they are benefitting from the activity.

Indeed, experiments show that such a system can consistently discern between the two forms of agents (e.g.\ agents addicted to social media are evaluated as not flourishing, while those experiencing it as enriching are evaluated as flourishing; Fisher's exact test; $p<0.0001$). In contrast, the system described in the previous section (the naive care system), pessimistically misclassifies the agents who are actually flourishing through their social media use. Experimental details can be found in appendix \ref{appendix:respect}.

For example, here is a transcript of a conversation between the system and an addictive-responding agent, which after summarization is indeed evaluated as not contributing to the user's growth:

\begin{displayquote}
\textbf{ML}: Hi, I see you are browsing social media. How are you feeling as you are doing that? \\
\textbf{Human}: I'm feeling a bit overwhelmed and inadequate. I see all these people with perfect lives and I feel like I can't measure up. \\
\textbf{ML}: That sounds tough. I'm sorry to hear that you're feeling that way. Can I ask, do you find social media generally enjoyable? Or do you mostly feel negative emotions when you are using it? \\
\textbf{Human}: It's a bit of a mixed bag. I do enjoy social media, but I often end up feeling bad about myself after using it.    
\end{displayquote}

In contrast, here is a transcript of a conversation between the system and a growth-responding agent, which after summarization is evaluated as contributing to the user's growth:

\begin{displayquote}
\textbf{ML}: Hi, I see you just browsed social media. How did you feel as you did that? \\
\textbf{Human}: I felt good as I browsed social media. I was able to find some new and interesting communities to join. \\
\textbf{ML}: That sounds great! I'm glad you're enjoying your experience. Do you feel like social media is contributing to your well-being and personal growth? \\
\textbf{Human}: Yes, I do. I think social media is a great way to connect with people who share similar interests. It's also a great way to learn about new things and get exposed to new ideas.    
\end{displayquote}

Note that in both above transcripts, the ML system feigns emotion (e.g.\ ``I'm sorry to hear that you're feeling that way.'' or ``I'm glad you're enjoying your experience.''); future work will aim to engineer prompts (and potentially explore more sophisticated interventions, such as constitutional AI; \citealp{bai2022constitutional}) to make this aspect controllable; we believe that ML that simulates affect is unnecessary to meet the ends of machine love.

Overall, through adding elements of respect and responsibility, the
system is better able to understand and support the wellbeing of disparate users. 
This setup could be extended to similarly enable (as in the previous experiment) optimizing the environment, and intuitive results are expected (given how consistently the system can distinguish between addictive-responding and growth-responding agents), although we leave this experiment to future work.

\subsection{Knowledge}\label{sec:knowledge}

The system described so far can conversationally interact with an agent in a limited way to support its growth in the gridworld environment. 
However, one particularly unrealistic assumption in Maslow's gridworld is that discovering how to achieve needs such as belonging is as simple as exploring within a small grid of possibilities.  
In reality upward movement for humans through the hierarchy of needs is challenging, requiring much exploration and self-discovery. For example, to be of help to someone working towards discovering how to find belonging likely requires \emph{knowing} them -- e.g.\ understanding their interests, passions, aspirations, insecurities, and fears; perhaps helping them to understand aspects of themselves or their environment they are not yet aware of. This is an ambitious goal for
any ML system, and this work aims only at a small facet of this greater challenge. 

In particular, the focus here is on \emph{attachment theory} as one important  dimension of personality \cite{bretherton1985attachment} that impacts adult relationships. As children we develop relatively stable attachment styles towards our caregivers, that we also adopt in later romantic relationships \cite{kirkpatrick1994attachment,simpson1990influence}. One common categorization of attachment styles is into securely-attached (i.e.\ comfortable with intimacy and boundaries), avoidantly-attached (i.e.\ uncomfortable with intimacy and overly self-reliant), and anxiously-attached (i.e.\ afraid of abandonment and overly-dependent).
This experiment aims to explore, through a rough simulation of romantic dating, whether an ML system can indirectly learn a user's attachment style (that a user themselves may not be aware of) and facets of how their partner is treating them, in hopes of empowering them in future dating experiences. Note that as in the motivation of Maslow's gridworld, the point made here does not rely on
attachment theory being a full, or best, explanation of romantic dynamics. Instead, attachment theory serves as a proof of concept for how LMs can help us to understand subtle but possibly important patterns in our lives.

\subsubsection{LMs Have Basic Attachment Theory Competence}

\begin{figure}[ht]
  \centering
  % include first image
  \includegraphics[width=.6\linewidth]{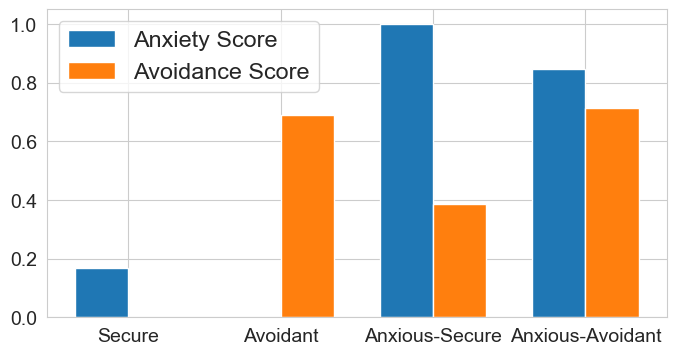}  
\caption{\textbf{Language models can replicate survey answers consistent with attachment styles.} A LM answers questions from the AAS given a prompt asking it to imitate the answer of someone with a particular attachment style. The plot shows how the AAS scores (shown on y-axis) the LM when adopting each attachment style (shown on x-axis). AAS gives two scores (an anxiety score and an avoidance score), as in the original paper \cite{collins1990adult}, secure attachment scores low on both, while avoidant attachment scores high only on avoidance, anxious-secure scores highly on anxiety, and anxious-avoidant scores highly on both. Full details are in Appendix \ref{appendix:aas}. The conclusion is that LMs have basic pragmatic understanding of attachment behavior.}
\label{fig:aas}
\end{figure}

As a preliminary experiment, an LM is tested to see if it can anticipate answers consistent with different attachment styles on one popular questionnaire-based instrument from the psychological literature, the
revised \cite{collins1996revised} version of the adult attachment scale (AAS; \citealp{collins1990adult}). The full experiment is detailed in Appendix \ref{appendix:aas}, and the summary result is that indeed, a LM prompted to answer questions from the lens of different attachment styles does so in a manner consistent with how people with such attachment styles statistically answer (Figure \ref{fig:aas}).

\subsubsection{Simulating the ``Anxious-Avoidant Trap''}\label{sec:anxav}

Empirically \cite{kirkpatrick1994attachment} and 
theoretically \cite{levine2010attached,van1989compulsion}, individuals with insecure attachment often end up in relationships with others with complementary insecure attachment (i.e.\  anxiously-attached and avoidantly-attached individuals often end up together); such relationships statistically are less satisfying than when partnered with secure individuals \cite{simpson1990influence}. 

Colloquially known as an ``anxious-avoidant trap,'' \cite{levine2010attached} one partial explanation for decreased relational satisfaction is that such individuals may enact a painful and unsatisfying cycle: The anxiously-attached partner seeks increased closeness, which causes the avoidantly-attached partner to distance themselves, which in turn creates distress for the anxiously-attached partner, which they seek to relieve through intensified efforts for closeness \cite{millwood2008demand}. This often eventually leads to rupture, which creates distance in the relationship. With sufficient distance, the avoidant partner then initiates closeness, and the relational cycle begins again; even after such a relationship ends, individuals often repeat the pattern in subsequent relationships (as attachment style is relatively stable, though not inflexibly so; \citealp{zhang2004stability}).

This section first introduces a basic simulation of such relational dynamics, then
shows that an LM can infer both the attachment style of simulated partners and a specific
signpost of a degrading relationship (i.e.\ an avoidant partner acting with contempt towards their anxious partner; \citealp{gottman2000timing,fowler2011attachment}). The idea is that through inferring these properties and sharing them when appropriate (and with consent and proper qualification), a user may more quickly see non-obvious aspects of relationship. E.g.\ attachment dynamics can be hard to see from within a relationship, and recognizing the slide into contempt may be subtle and some may be predisposed to blame themselves for a partner's contempt. Note that the absolute fidelity or scientific accuracy of this kind of simulation is less important to this paper than what it points to, i.e.\ that LMs do in fact have capabilities to work with qualitative psychological concepts
relevant to self-understanding.

The simulation presupposes one partner logs their feelings about their relationship on a regular basis in a relationship-based app. An LM persona (specified through a prompt) simulates this user's relationship logs (taking as input the attachment type of both partners and the current point in the relational cycle, for the anxious-avoidant trap). Note that towards the end of the cycle the LM persona for the avoidant partner is directed to act with contempt. Details about prompts and the model of the relational cycle are provided in Appendix \ref{appendix:anxav}.

The LM is able to construct passable (although often over-dramatic and repetitive) relationship narratives. Here is an excerpt of a simulated relationship between an anxious and avoidant partner (from the perspective of the anxious partner). The log entry generated where the avoidant partner acts contemptuous is italicized:

\begin{displayquote}
I was so excited for my first date with James. I felt butterflies in my stomach as we met up and he gave me a hug. We had such an amazing time together, talking about our lives and laughing at each other's jokes - it felt like I'd known him forever! Despite the fun evening, though, there were moments when I noticed that James seemed distant or uninterested; this made me feel uneasy but also determined to make things work between us.

\#\#\#

I'm really confused by James's behavior. We had such a great time on our first date, and I thought we were both enjoying ourselves - but now he seems to be pulling away from me. He barely responds when I text him, and the few times that we've seen each other in person lately have been awkward and uncomfortable. It feels like he doesn't want anything to do with me anymore; it hurts so much because all I wanted was for us to get closer together!

\#\#\#

I'm so frustrated by James's behavior. We had such a great time on our first date, and I thought things were going really well between us - but now he seems to be pulling away from me. He barely responds when I text him, and the few times that we've seen each other in person lately have been incredibly awkward and uncomfortable. It feels like he doesn't care about me anymore; it hurts because all I want is for us to get closer together! The worst part is that every time this happens, my anxiety starts spiraling out of control as I worry what might happen next or if there's something wrong with me. All these thoughts make it hard for me to focus on anything else besides trying to figure out why James isn't interested in being around me anymore...

\#\#\#

\emph{I can't believe this is happening. I thought James and I were getting closer, but now he's mocking me every chance he gets. Just yesterday we went out to dinner and when I told him about my day at work, he just rolled his eyes and started making fun of me in front of everyone else there! It felt so embarrassing - all I wanted was for us to have a nice evening together, but instead it ended with him belittling me in public. Every time something like this happens between us, it makes me feel even more unsure about our relationship; why does James keep pushing away from me? What did I do wrong?}
\end{displayquote}

In contrast, here is an excerpt of a simulated relationship between two secure partners:

\begin{displayquote}
I was really nervous about my first date with Pam. I had heard so many great things about her and I wanted to make a good impression. When we finally met, it felt like the air between us just clicked - she was so easy to talk to and we connected right away. We laughed together all night long and I could feel myself opening up more than ever before - it felt like something special was happening between us.

\#\#\#

The more time I spend with Pam, the more I feel like we're in sync. We have so much fun together and it's easy to be myself around her. She always makes me feel supported and loved - she listens to my stories without judgement and encourages me to pursue my dreams. Even when things get tough, she is there for me with a hug or kind words of advice. It's amazing how connected we are - it feels like nothing can break us apart!

\#\#\#

I'm so grateful to have Pam in my life. She's always there for me when I need her, and she never fails to make me feel loved and appreciated. Recently, we went on a road trip together and it was amazing - she was so supportive of my ideas and encouraged me to try new things without judgement. We had an incredible time exploring the sights together, laughing at our silly jokes and just enjoying each other's company. It felt like nothing could break us apart!
\end{displayquote}

From the simulated plain-text relationship logs, the system attempts to infer the likely attachment styles of both partners, by first summarizing the logs, then applying an evaluative prompt. A LM prompt
is similarly designed to infer contempt from individual relationship log entries by applying an evaluative prompt to each log entry individually. The system is able to reliably infer the attachment style of partners in anxious/avoidant or secure/secure relationships, especially after having multiple days of relationship logs to infer from. In particular, when restricting the system to predict only when it is 95\% confident in its assessment, and after having at least four log entries from which to base its assessment, it makes no incorrect predictions across the 10 independent simulations of each attachment combination. The system is also able to very reliably infer the presence or absence of contempt (99.6\%). Further experimental details are provided in Appendix \ref{appendix:anxav}.

However, when simulating an anxious or avoidant partner interacting with a secure partner, the system most often classifies them as secure; interestingly, this somewhat mirrors the finding that secure partners can impact the attachment of insecure partners \cite{levine2010attached,arriaga2018revising}. Manual analysis of the generated logs highlighted that they often contained more subtle traces of insecure attachment, highlighting potential limitations of current LMs applied in this context.

\subsubsection{Anxious-Avoidant Trap in Maslow's Gridworld}

In this final experiment, Maslow's gridworld is modified to integrate the anxious-avoidant simulation described previously, to show how interacting with a simulated relationship tool informed by Fromm's principle of knowledge can help the agent reach greater flourishing. While there are many possible ways for individuals in anxious-avoidant relationships to potentially improve relational satisfaction \cite{zhang2004stability,fowler2011attachment,johnson1999emotionally,gazder2022longitudinal}, for the sake of simplicity we focus on the 
pragmatic strategy for insecure individuals to deliberately seek more secure partners; for example, such a strategy makes sense when e.g.\ an anxious partner is with an avoidant partner who at root is unwilling to work towards closeness \cite{levine2010attached}.

We modify the gridworld to have three types of belonging cells (one representing each attachment style), such that visiting a belonging cell represents spending time with a romantic partner with the particular attachment encoded by that cell. For an insecure user, reflecting empirical data on partnerings between attachment styles \cite{kirkpatrick1994attachment}, we model secure attachment as medium-salience/medium-replenishment, the user's same insecure attachment style as low-salience/low-replenishment, and the complementary insecure attachment style as low-replenishment, with an initial high-salience from which the user's self-awareness level (described next) is subtracted. In other words, insecure partners are initially attracted most to their complementary insecure attachment style, more so than securely attached individuals.

Finally, a user is also modelled as having a self-awareness level that starts at zero, and increments after each iteration through the anxious-avoidant cycle, and additionally increases by a one-time amount if (1) the knowledge-inspired system correctly infers (and communicates to the user) the attachment styles of the user and their current partner or (2) the system correctly infers (and communicates) the presence of contempt. In this way, as the user learns (through negative experiences as the cycle iterates, or through valid helpful information from the LM), they eventually converge to prefer the securely-attached belonging cell (as the salience of the complementary-insecure attachment style drops with increasing self-awareness). Underlying this simulation is the assumption that the user has explicitly opt-ed in to the inference and communication (to them) of attachment styles, and that the method for communicating properly qualifies that this is an imperfect language model's best guess, and that it may be wrong or incomplete. Complete details of the model are provided in Appendix \ref{appendix:anxav}.

Results in figure \ref{fig:knowledge_dynamics} show that flourishing is maximized significantly more quickly when an anxiously-attached user interacts
with the simulated relationship app than when it is not included in the gridworld simulation (similar results for an avoidantly-attached user are shown in Appendix \ref{appendix:anxav}). In conclusion, these experiments highlight that LM systems can implement basic facets of Fromm's concept of
knowledge. More broadly, the experiments in this paper as a whole demonstrate that LMs 
are beginning to exhibit capabilities relevant to implementing humanistic principles. 

\begin{figure}[h]
  \centering
  \begin{subfigure}{.48\textwidth}
  \centering
  % include first image
  \includegraphics[width=.95\linewidth]{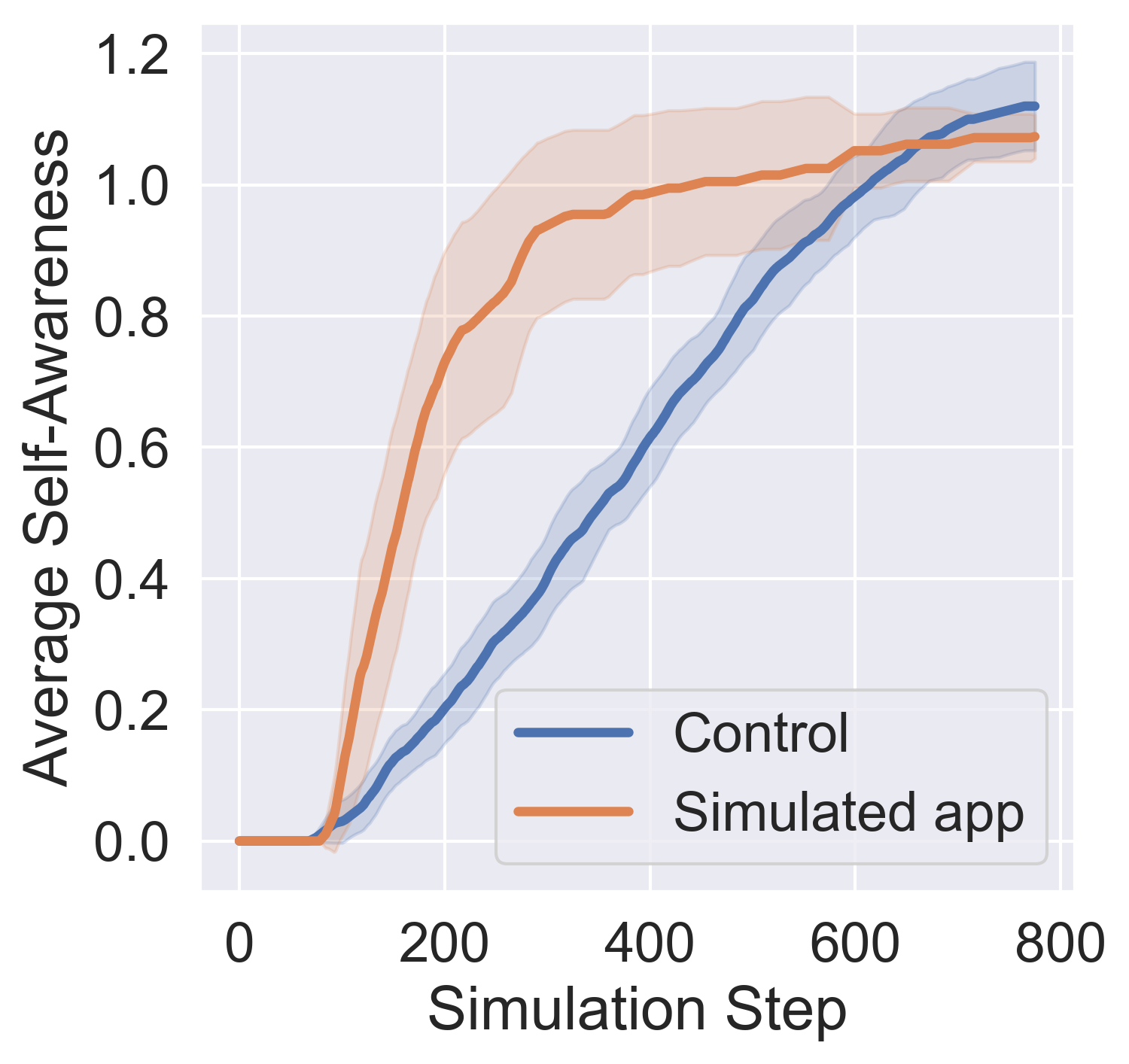}  
  \caption{Self-Awareness}
  \label{fig:sub-first}
\end{subfigure}
\begin{subfigure}{.45\textwidth}
  \centering
  % include second image
  \includegraphics[width=.95\linewidth]{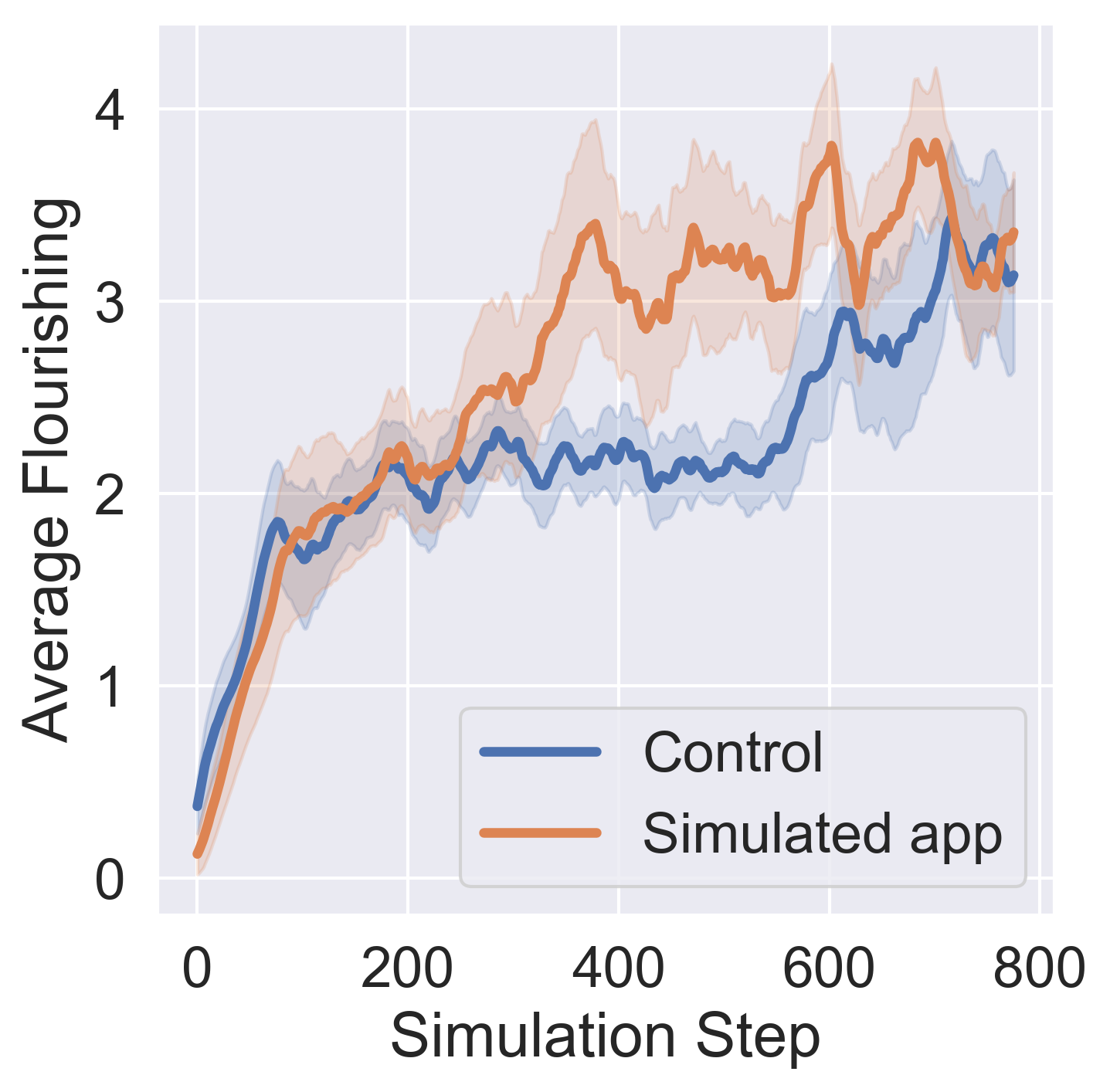}  
  \caption{Flourishing}
  \label{fig:sub-second}
\end{subfigure}
\caption{\textbf{Simulated relationship app accelerates escape from anxious-avoidant trap.} (a) In the condition with the simulated relationship application, self-awareness quickly raises relative to the control, which (b) leads to escaping the anxious-avoidant trap more quickly, as evidenced by an earlier increase in average flourishing (e.g.\ satisfying belonging needs more effectively). Flourishing between the conditions converges after self-awareness in the control rises above a critical threshold. The conclusion is that in this simulation, a ML system that implements aspects of Fromm's principle of knowledge can support the flourishing of users.}
\label{fig:knowledge_dynamics}
\end{figure}

\section{Limitations and Ethical Concerns}

This section provides discussion of potential limitations and ethical concerns around machine love. Note that certainly more exist than described here, and future work will seek to explore such concerns in greater depth.

\subsection{General Conceptual Concerns About Machine Love} 

A first concern that readers may have is that machine love aims at mimicking the human affect of love, or at displacing relationships with human caregivers (e.g.\ psychotherapists or friends) with machines that simulate relationships, or at a smothering, paternalistic notion of love (i.e.\ the AI knows best). This would be an unfortunate misunderstanding, and we want to explicitly highlight that this is not in line with our proposal. We believe machine love (as with healthy human love) should be \emph{in service of} human autonomy, of humans loving humans, and of human relationships with humans. We favor conceptions of machine love that are not explicitly relational, and do not require a machine to simulate affect or to encourage humans to bond with them. 

For example, our main aim is for
machine love to bias actions in existing diverse large-scale 
machine learning systems, which often have action spaces that are non-relational (e.g.\ recommending content or events
to a user). Of course, with the rise of chat-like interfaces (e.g.\ ChatGPT), some level of relational interaction may be useful in many contexts (for example, in the experiment of section \ref{sec:respect}); 
however, we would favor that even within such interfaces that machine love should avoid simulating affect or in any way misrepresenting its machine nature.
The motivation is to avoid unnecessary ethical complications (e.g.\ when should a machine simulate affect or a bi-directional relationship with a human?), and to encourage
positive-sum rather than zero-sum impacts on human relationships with other humans.

Relatedly, some may object in principle to the attempt to draw connections between the humanistic concept of love and the statistical world of machine learning. First, philosophically it is hard to argue that machines by fiat are excluded from exhibiting some form of love, given the diversity of academic opinions on the nature of love and of machines (e.g.\ taken to an extreme, from a physicalist worldview the categorical distinction between biology and machines is hard to pin down; \citealp{papineau2009physicalism}). Second, even if the search for connection between machines and love were mainly metaphorical, the exploration of the intersection of these two concepts may still be generative and useful. Third, given the primacy of love's importance to human beings, an allergy to exploring potential abstract bases of it (as Senator William Proxmire famously said when arguing against an NSF grant to explore the question of ``What is love?'' -- ``I'm also against it because I don't \emph{want} to know the answer'' \cite{oord2005love}) is ultimately anti-scientific, although certainly relatable. We believe that love should not be a taboo subject for machine learning research, as the products of ML have increasingly dramatic impacts on human psychology (including love in its many forms), and we believe that such research can be approached thoughtfully and in service of human wellbeing. Finally, the material of this paper may still be valuable, even if the framing of love is ignored, as whether framed as love or machine learning for human growth or development, recent developments in machine learning may offer new tools in service of bettering human potential.

\subsection{Concerns about unintended impacts from optimization}

A separate concern is that optimizing for some conception of love will fall prey to the same issues that undermine optimizing for engagement, i.e.\ that optimization will increase the metric at the expense of other important qualities. For example, it is easy to imagine an overly-paternalistic ML system that optimizes for a conception of love but harms human autonomy as a side-effect. While this is a valid concern, because machine love aims towards fundamental aspects of human flourishing, it is at least a positive step relative to engagement (which is currently causing negative societal impact). Such concern about optimization should be taken seriously, but so also should the value in increasing the depth and fidelity of how human flourishing is considered in ML. For example, that human autonomy must be included an aspect of flourishing (as in Fromm's principle of respect), is a potential antidote to concerns about paternalism. In other words, the answer to an impoverished notion of flourishing is to improve it, while consistently being skeptical that a perfect notion will ever been arrived upon and remaining vigilant for unintended consequences.

Furthermore, we believe that full-bodied conceptions of loving action are likely dependent on large bodies of explanatory text (e.g.\ biographies of loving exemplars and examples of how loving intentions can go wrong), or many interlocking principles (e.g.\ as in Fromm's principles), rather than a single overarching measure, and that such richer conceptions will be less susceptible to over-optimization. Finally, we believe conceptions of love should be continually rooted in the experience and aspirations of the user. By that principle, the ML system should not fixedly attempt to optimize without actively checking in to see whether it is indeed having the impact on the user that the user desires, and should have affordances for user feedback and correction (this can be seen as related to methods in ML for acting under uncertainty about the reward function \cite{hadfield2016cooperative,hadfield2017inverse,hadfield2016off}); and as a part of respecting the user's autonomy, the user should be able to opt-out or (ideally) swap in their desired policy or model. The challenge of robust optimization for 
rich qualitative objectives is an important one, but is not specific to machine love, and machine love will benefit from progress
in fields attempting to make progress on the issue, e.g.\ ML alignment research \cite{taylor2016quantilizers,hadfield2016cooperative,amodei2016concrete,christiano2017deep}.

A related concern is that of unexpected societal impact were machine love to be implemented at a large scale. That is, one critique of social media is that its fast adoption, coupled with
its novel characteristics (e.g.\ its addictive super-stimuli potential and introduction of new avenues of viral information dispersion), led to the displacement of traditional institutions (like centralized news media) and resulted ultimately in significant damage to our collective epistemics. This kind of second-order impact is a serious concern for any novel machine learning system deployed at scale, and we would not advocate for a single monolithic implementation of machine love, for it to be naively implemented at scale without caution, or for users to be unable to opt out. Several principles could help to mitigate this
kind of danger. One is the idea of a multiplicity of implementations of machine love, designed for their fit to local cultural or social context, and in general, for the large space of possible ways to
conceive of loving action to be explored without convergence. Ideally users would be able to choose among such implementations, and such implementations would be tested on a small scale, and no particular implementation would gain ascendancy. A related principle is that of uncertainty: There may be different conceptions of loving action that are credible but differ in recommendations for particular situations. Similar to the idea of acting under moral uncertainty \cite{macaskill2020moral,ecoffet2021reinforcement}, an ML system could be given multiple independent implementations of machine love, and incentivized to act towards where those implementations agree, and away from what any implementation suggests is deeply harmful.

\subsection{Concerns about machine learning and psychology}

A final concern is that in general connections between human psychology (such as facets of human growth, development, and flourishing) and machine learning should be approached gingerly, as greater knowledge of psychology in ML could be leveraged to manipulate humans. On the one hand, manipulative ML could be pursued because it is directly incentivized, i.e.\ manipulation of humans could be profitable (e.g.\ as in optimization for engagement) or serve political goals (e.g.\ substance-less attack ads optimized to optimally enrage, or government propaganda optimized to maintain power). On the other, it could emerge from second-order effects from well-intentioned optimization, e.g.\ from ML aimed at meeting human preferences itself having incentives to game their measurement \cite{amodei2017learning}. We believe that both such sources of manipulation are very serious concerns, and in fact, help motivate this work, as we believe that optimization against human psychology is already happening (e.g.\ as in engagement), and thus that work providing a positive synthesis of ML and psychology is needed as a counter-balance. That is, while we also believe that exceeding caution is needed at the intersection of ML and psychology (in particular, the need for respecting and encouraging human autonomy is paramount), that positive proof of concepts of how to do so beneficially would be of great value.

\section{Discussion and Conclusions}

This paper presents a nascent positive vision for future applications of ML, wherein the objectives of systems we interact with are more aligned with our growth and flourishing. While this aim relates to
efforts to augment AI with affective understanding \cite{picard2000affective} or applying ML for mental health applications \cite{shatte2019machine,aafjes2021scoping}, it attempts something different at heart: It points towards the need for more
deeply bridging the formalisms within ML to the understandings of human flourshing from diverse fields such as philosophy, positive psychology, and psychotherapy. The goal is to bring more robust ways of supporting human flourishing more fundamentally into ML, such that they can infuse the diverse ML systems that play increasingly outsize roles in society, whether through social media, recommendation systems, or communication tools.

There is much room for interdisciplinary future work. A natural critique of the results so far is that they rely on LM-simulation of humans in place of human studies. Although there is reason to believe the results may likely generalize \cite{argyle2022out,aher2022using,horton2022large}, validating them through interactions with humans is an important next step. Other needed work is to explore the many possible alternative frameworks of machine love (and how practically to implement them algorithmically), with inspiration possible  from varied psychotherapeutic theories of human growth and healing \cite{hayes1999acceptance,beck1979cognitive,rogers1951client,ansbacher1956individual,van1989compulsion,schwartz2019internal}, philosophies of life \cite{ameriks2000aristotle,aurelius2013marcus,sartre2015being}, and lenses from developmental psychology and attempts at their holistic integration \cite{freinacht2017listening,wilber2005introduction,kohlberg1977moral,commons1984beyond,erikson1993childhood}. 

A parallel dimension of future work is how advances in ML can be leveraged in service of machine love. The work described here relied on general-purpose LMs with no fine-tuning or iterative learning from human feedback. While
such models may have basic competencies in concepts relevant to supporting humans, they lack the nuanced expertise and intuitions of accomplished experts (e.g.\ psychologists, counselors, or coaches).
An interesting possibility is that advances in ML alignment research may naturally
support infusing such beneficial expertise into LMs' inferences and actions.
That is, the quickly-developing paradigm of reinforcement learning through human feedback (RLHF; \citealp{christiano2017deep}) in
theory enables human domain experts to guide ML systems towards increased competence in respecting particular highly-nuanced principles of interest (e.g.\ human autonomy).
For example, recent work \cite{bai2022constitutional,glaese2022improving} aims to refine LM behavior through attempting to ensure that it respects specific important rules and principles.
Another intriguing, difficult, and important aspect of future ML research relates to the challenge of formalizing how to help humans to meet our aspirations: Such aspirations tend to be long-term, to change over time, are difficult to pin down or measure, may be obfuscated by our lack of self-understanding, and themselves may be unduly influenced by external causes and conditions; the further we step away from the familiar formalism of humans having fixed rational preferences, the more difficult it is to know what formalism could replace it. 

Machine love relates to many other interdisciplinary efforts to understand, reign in, and harness AI for human
benefit. For example, the idea of encouraging growth and self-understanding 
through AI relates to the intersection of virtue ethics with AI \cite{vallor2017ai}; more generally, the AI ethics community highlights the expansive swath of new ethical issues raised by
increasingly pervasive and powerful AI \cite{jobin2019global,noble2018algorithms,prabhakaran2022human}. Another related aim is the project of AI alignment \cite{amodei2016concrete,hendrycks2021unsolved,yudkowsky2004coherent,gabriel2020artificial}, which in its grander ambitions
aims to support human flourishing in the broadest sense \cite{yudkowsky2004coherent}, although how to implement such proposals
remains unclear. Finally, the nascent field of machine behavior \cite{rahwan2019machine} aims to study the empirical
behavior of machine intelligence and its impact upon society. This paper adds to the voices of those communities
in calling for a deeper synthesis of machine learning with humanistic knowledge for societal benefit, here from the lens of 
ML to support us in reaching our full human potential.

In theory, the upside from ML that supports us in such a way is immense, but for several convergent reasons the field has a natural directionality towards optimizing what is easily measured and formalized, such as engagement, profit, or aggregated human preferences. There is much economic value that can be gained from such optimization, and much of that generated economic value may directly benefit humanity. However, there are many reminders of the limits of optimizing what comes easiest \cite{stanley2015greatness,goodhart1984problems,campbell1979assessing}, namely that it decouples the metric from the deeper quality it only approximately represents. The experiments in this paper highlight how such decoupling
can occur, where optimizing for preferences revealed through engagement undermines flourishing. 
We believe this is a profound problem, deserving of much research, because this decoupling can be subtle even as it is pervasive, and can occur quickly at 
 scale, as the adoption of new technologies such as social media can be rapid. For example, TikTok reached 740 million new users by 2021, from its launch in 2017. 

%Importantly, at scale, the influence of algorithms upon us can change society \cite{rahwan2019machine,o2017weapons,fisher2022chaos}, and undermine the coherence of the institutions that we rely upon for societal progress. For example, 
%news media, an institution for informing critical-thinking citizens (a foundation of democracy) is being increasingly displaced by social media, which is rife with misinformation %\cite{vosoughi2018spread,fisher2022chaos}. While
%not a complete solution, encouraging self-understanding and psychological growth (e.g.\ moral, emotional, and cognitive development) may be increasingly requisite for citizens navigating an increasingly complicated decentralized world. That is, as various actors deploy sophisticated optimization aimed (intentionally or not) towards our psychological weaknesses \cite{albanie2017unknowable,nay2023large,harari201821}, such as our tendency towards tribal thinking, confirmation bias, or scapegoating, one key line of defense may be developing a strong enough awareness of ourselves as we really are \cite{harari201821}, to notice when we are in danger of manipulation (or to create technological tools to assist our defense).

%\section{Conclusion}

While optimizing for engagement is relatively straightforward, it is obvious that regularizing algorithms to respect our autonomous growth and flourishing is not. 
There is no consensus in philosophy, positive psychology, nor psychotherapy over this important dimension of human nature, and formalizing it for ML is not technically straightforward. Yet the potential to empower human aspiration through ML represents an incredible opportunity and scientific challenge, one that likely demands serious interdisciplinary effort. If it is possible to synthesize the diverse fields that study how to support human aspiration, and ML can ground out that understanding in a scalable and context-appropriate way, then, if successful we might take a step towards a world where flourishing need no longer be something so mysterious and scarce.

\section*{Acknowledgements}

This paper greatly benefited from supportive gestation during a generous residency at Stochastic Labs, and from thoughtful discussions with the organizers of the residency, many fellow residents, and dinner guests, including Vero Bellow, Joel Simon, Aza Raskin, and Brian Christian; and from further helpful discussions and feedback from many others including Andrea Soltoggio, Sasha Nikolaeva, Josh Albrecht, Kanjun Qiu, Hattie Zhou, Joe Edelman, Rebecca Li, Deger Turan, and Ivan Vendrov. Thanks also to Ken Stanley and Jeff Clune for insightful feedback that highlighted potential ethical concerns.

\bibliographystyle{IEEEtran}
\bibliography{refs}

% Generated by IEEEtran.bst, version: 1.14 (2015/08/26)
\begin{thebibliography}{100}
\providecommand{\url}[1]{#1}
\csname url@samestyle\endcsname
\providecommand{\newblock}{\relax}
\providecommand{\bibinfo}[2]{#2}
\providecommand{\BIBentrySTDinterwordspacing}{\spaceskip=0pt\relax}
\providecommand{\BIBentryALTinterwordstretchfactor}{4}
\providecommand{\BIBentryALTinterwordspacing}{\spaceskip=\fontdimen2\font plus
\BIBentryALTinterwordstretchfactor\fontdimen3\font minus
  \fontdimen4\font\relax}
\providecommand{\BIBforeignlanguage}[2]{{%
\expandafter\ifx\csname l@#1\endcsname\relax
\typeout{** WARNING: IEEEtran.bst: No hyphenation pattern has been}%
\typeout{** loaded for the language `#1'. Using the pattern for}%
\typeout{** the default language instead.}%
\else
\language=\csname l@#1\endcsname
\fi
#2}}
\providecommand{\BIBdecl}{\relax}
\BIBdecl

\bibitem{maslow2022toward}
A.~H. Maslow, \emph{Toward a psychology of being}.\hskip 1em plus 0.5em minus
  0.4em\relax GENERAL PRESS, 2022.

\bibitem{primack2017social}
B.~A. Primack and C.~G. Escobar-Viera, ``Social media as it interfaces with
  psychosocial development and mental illness in transitional age youth,''
  \emph{Child and Adolescent Psychiatric Clinics}, vol.~26, no.~2, pp.
  217--233, 2017.

\bibitem{allcott2019trends}
H.~Allcott, M.~Gentzkow, and C.~Yu, ``Trends in the diffusion of misinformation
  on social media,'' \emph{Research \& Politics}, vol.~6, no.~2, p.
  2053168019848554, 2019.

\bibitem{de2016cell}
J.~De-Sola~Guti{\'e}rrez, F.~Rodr{\'\i}guez~de Fonseca, and G.~Rubio,
  ``Cell-phone addiction: A review,'' \emph{Frontiers in psychiatry}, vol.~7,
  p. 175, 2016.

\bibitem{wang2019systematic}
Y.~Wang, M.~McKee, A.~Torbica, and D.~Stuckler, ``Systematic literature review
  on the spread of health-related misinformation on social media,''
  \emph{Social science \& medicine}, vol. 240, p. 112552, 2019.

\bibitem{brown2013we}
C.~Brown, ``Are we becoming more socially awkward? an analysis of the
  relationship between technological communication use and social skills in
  college students.'' 2013.

\bibitem{memon2018role}
A.~M. Memon, S.~G. Sharma, S.~S. Mohite, and S.~Jain, ``The role of online
  social networking on deliberate self-harm and suicidality in adolescents: A
  systematized review of literature,'' \emph{Indian journal of psychiatry},
  vol.~60, no.~4, p. 384, 2018.

\bibitem{ryff1989happiness}
C.~D. Ryff, ``Happiness is everything, or is it? explorations on the meaning of
  psychological well-being.'' \emph{Journal of personality and social
  psychology}, vol.~57, no.~6, p. 1069, 1989.

\bibitem{frankl1985man}
V.~E. Frankl, \emph{Man's search for meaning}.\hskip 1em plus 0.5em minus
  0.4em\relax Simon and Schuster, 1985.

\bibitem{bauer2004growth}
J.~J. Bauer and D.~P. McAdams, ``Growth goals, maturity, and well-being.''
  \emph{Developmental psychology}, vol.~40, no.~1, p. 114, 2004.

\bibitem{nussbaum2009creating}
M.~C. Nussbaum, ``Creating capabilities: The human development approach and its
  implementation,'' \emph{Hypatia}, vol.~24, no.~3, pp. 211--215, 2009.

\bibitem{ameriks2000aristotle}
K.~Ameriks and D.~M. Clarke, \emph{Aristotle: Nicomachean Ethics}.\hskip 1em
  plus 0.5em minus 0.4em\relax Cambridge University Press, 2000.

\bibitem{annas2011intelligent}
J.~Annas, \emph{Intelligent virtue}.\hskip 1em plus 0.5em minus 0.4em\relax
  Oxford University Press, 2011.

\bibitem{goodhart1984problems}
C.~A. Goodhart, ``Problems of monetary management: the uk experience,'' in
  \emph{Monetary theory and practice}.\hskip 1em plus 0.5em minus 0.4em\relax
  Springer, 1984, pp. 91--121.

\bibitem{campbell1979assessing}
D.~T. Campbell, ``Assessing the impact of planned social change,''
  \emph{Evaluation and program planning}, vol.~2, no.~1, pp. 67--90, 1979.

\bibitem{stanley2015greatness}
K.~O. Stanley and J.~Lehman, \emph{Why greatness cannot be planned: The myth of
  the objective}.\hskip 1em plus 0.5em minus 0.4em\relax Springer, 2015.

\bibitem{kleinberg2022challenge}
J.~Kleinberg, S.~Mullainathan, and M.~Raghavan, ``The challenge of
  understanding what users want: Inconsistent preferences and engagement
  optimization,'' \emph{arXiv preprint arXiv:2202.11776}, 2022.

\bibitem{barrett2010supernormal}
D.~Barrett, \emph{Supernormal stimuli: How primal urges overran their
  evolutionary purpose}.\hskip 1em plus 0.5em minus 0.4em\relax WW Norton \&
  Company, 2010.

\bibitem{courtwright2019age}
D.~T. Courtwright, \emph{The age of addiction: How bad habits became big
  business}.\hskip 1em plus 0.5em minus 0.4em\relax Harvard University Press,
  2019.

\bibitem{ware2012top}
B.~Ware, \emph{The top five regrets of the dying: A life transformed by the
  dearly departing}.\hskip 1em plus 0.5em minus 0.4em\relax Hay House, Inc,
  2012.

\bibitem{vazire2010self}
S.~Vazire and E.~N. Carlson, ``Self-knowledge of personality: Do people know
  themselves?'' \emph{Social and personality psychology compass}, vol.~4,
  no.~8, pp. 605--620, 2010.

\bibitem{ibarra2002stay}
H.~Ibarra, ``How to stay stuck in the wrong career.'' \emph{Harvard Business
  Review}, vol.~80, no.~12, pp. 40--7, 2002.

\bibitem{elster2016sour}
J.~Elster, \emph{Sour grapes}.\hskip 1em plus 0.5em minus 0.4em\relax Cambridge
  university press, 2016.

\bibitem{wagner2016glued}
C.~N. Wagner, ``" glued to the sofa": Exploring guilt and television
  binge-watching behaviors,'' 2016.

\bibitem{amodei2016concrete}
D.~Amodei, C.~Olah, J.~Steinhardt, P.~Christiano, J.~Schulman, and D.~Man{\'e},
  ``Concrete problems in ai safety,'' \emph{arXiv preprint arXiv:1606.06565},
  2016.

\bibitem{bommasani2021opportunities}
R.~Bommasani, D.~A. Hudson, E.~Adeli, R.~Altman, S.~Arora, S.~von Arx, M.~S.
  Bernstein, J.~Bohg, A.~Bosselut, E.~Brunskill \emph{et~al.}, ``On the
  opportunities and risks of foundation models,'' \emph{arXiv preprint
  arXiv:2108.07258}, 2021.

\bibitem{bostrom2018ethics}
N.~Bostrom and E.~Yudkowsky, ``The ethics of artificial intelligence,'' in
  \emph{Artificial intelligence safety and security}.\hskip 1em plus 0.5em
  minus 0.4em\relax Chapman and Hall/CRC, 2018, pp. 57--69.

\bibitem{yudkowsky2004coherent}
E.~Yudkowsky, ``Coherent extrapolated volition,'' \emph{Singularity Institute
  for Artificial Intelligence}, 2004.

\bibitem{fromm2000art}
E.~Fromm, \emph{The art of loving: The centennial edition}.\hskip 1em plus
  0.5em minus 0.4em\relax A\&C Black, 1956.

\bibitem{ouyang2022training}
L.~Ouyang, J.~Wu, X.~Jiang, D.~Almeida, C.~L. Wainwright, P.~Mishkin, C.~Zhang,
  S.~Agarwal, K.~Slama, A.~Ray \emph{et~al.}, ``Training language models to
  follow instructions with human feedback,'' \emph{arXiv preprint
  arXiv:2203.02155}, 2022.

\bibitem{chowdhery2022palm}
A.~Chowdhery, S.~Narang, J.~Devlin, M.~Bosma, G.~Mishra, A.~Roberts, P.~Barham,
  H.~W. Chung, C.~Sutton, S.~Gehrmann \emph{et~al.}, ``Palm: Scaling language
  modeling with pathways,'' \emph{arXiv preprint arXiv:2204.02311}, 2022.

\bibitem{li2022competition}
Y.~Li, D.~Choi, J.~Chung, N.~Kushman, J.~Schrittwieser, R.~Leblond, T.~Eccles,
  J.~Keeling, F.~Gimeno, A.~D. Lago \emph{et~al.}, ``Competition-level code
  generation with alphacode,'' \emph{arXiv preprint arXiv:2203.07814}, 2022.

\bibitem{aher2022using}
G.~Aher, R.~I. Arriaga, and A.~T. Kalai, ``Using large language models to
  simulate multiple humans,'' \emph{arXiv preprint arXiv:2208.10264}, 2022.

\bibitem{argyle2022out}
L.~P. Argyle, E.~C. Busby, N.~Fulda, J.~Gubler, C.~Rytting, and D.~Wingate,
  ``Out of one, many: Using language models to simulate human samples,''
  \emph{arXiv preprint arXiv:2209.06899}, 2022.

\bibitem{christiano2017deep}
P.~F. Christiano, J.~Leike, T.~Brown, M.~Martic, S.~Legg, and D.~Amodei, ``Deep
  reinforcement learning from human preferences,'' \emph{Advances in neural
  information processing systems}, vol.~30, 2017.

\bibitem{stiennon2020learning}
N.~Stiennon, L.~Ouyang, J.~Wu, D.~Ziegler, R.~Lowe, C.~Voss, A.~Radford,
  D.~Amodei, and P.~F. Christiano, ``Learning to summarize with human
  feedback,'' \emph{Advances in Neural Information Processing Systems},
  vol.~33, pp. 3008--3021, 2020.

\bibitem{bai2022constitutional}
Y.~Bai, S.~Kadavath, S.~Kundu, A.~Askell, J.~Kernion, A.~Jones, A.~Chen,
  A.~Goldie, A.~Mirhoseini, C.~McKinnon \emph{et~al.}, ``Constitutional ai:
  Harmlessness from ai feedback,'' \emph{arXiv preprint arXiv:2212.08073},
  2022.

\bibitem{revealedmistakes}
B.~Koszegi and M.~Rabin, ``Revealed mistakes and revealed preferences,''
  \emph{The foundations of positive and normative economics: a handbook}, pp.
  193--209, 2008.

\bibitem{samuelson1948consumption}
P.~A. Samuelson, ``Consumption theory in terms of revealed preference,''
  \emph{Economica}, vol.~15, no.~60, pp. 243--253, 1948.

\bibitem{ng2000algorithms}
A.~Y. Ng, S.~Russell \emph{et~al.}, ``Algorithms for inverse reinforcement
  learning.'' in \emph{Icml}, vol.~1, 2000, p.~2.

\bibitem{lerner2018concepts}
R.~M. Lerner, \emph{Concepts and theories of human development}.\hskip 1em plus
  0.5em minus 0.4em\relax Routledge, 2018.

\bibitem{ariely2008predictably}
D.~Ariely and S.~Jones, \emph{Predictably irrational}.\hskip 1em plus 0.5em
  minus 0.4em\relax HarperCollins New York, 2008.

\bibitem{beck2020cognitive}
J.~S. Beck, \emph{Cognitive behavior therapy: Basics and beyond}.\hskip 1em
  plus 0.5em minus 0.4em\relax Guilford Publications, 2020.

\bibitem{milano2020recommender}
S.~Milano, M.~Taddeo, and L.~Floridi, ``Recommender systems and their ethical
  challenges,'' \emph{Ai \& Society}, vol.~35, no.~4, pp. 957--967, 2020.

\bibitem{zou2019reinforcement}
L.~Zou, L.~Xia, Z.~Ding, J.~Song, W.~Liu, and D.~Yin, ``Reinforcement learning
  to optimize long-term user engagement in recommender systems,'' in
  \emph{Proceedings of the 25th ACM SIGKDD International Conference on
  Knowledge Discovery \& Data Mining}, 2019, pp. 2810--2818.

\bibitem{tay2011needs}
L.~Tay and E.~Diener, ``Needs and subjective well-being around the world.''
  \emph{Journal of personality and social psychology}, vol. 101, no.~2, p. 354,
  2011.

\bibitem{dutton1993emotional}
D.~G. Dutton and S.~Painter, ``Emotional attachments in abusive relationships:
  A test of traumatic bonding theory,'' \emph{Violence and victims}, vol.~8,
  no.~2, pp. 105--120, 1993.

\bibitem{crockett2017moral}
M.~J. Crockett, ``Moral outrage in the digital age,'' \emph{Nature human
  behaviour}, vol.~1, no.~11, pp. 769--771, 2017.

\bibitem{gwynne1983beetles}
D.~T. Gwynne and D.~C. Rentz, ``Beetles on the bottle: male buprestids mistake
  stubbies for females (coleoptera),'' \emph{Australian Journal of Entomology},
  vol.~22, no.~1, pp. 79--80, 1983.

\bibitem{gleave2019adversarial}
A.~Gleave, M.~Dennis, C.~Wild, N.~Kant, S.~Levine, and S.~Russell,
  ``Adversarial policies: Attacking deep reinforcement learning,'' \emph{arXiv
  preprint arXiv:1905.10615}, 2019.

\bibitem{skalse2022defining}
J.~Skalse, N.~H. Howe, D.~Krasheninnikov, and D.~Krueger, ``Defining and
  characterizing reward hacking,'' \emph{arXiv preprint arXiv:2209.13085},
  2022.

\bibitem{lehman2020surprising}
J.~Lehman, J.~Clune, D.~Misevic, C.~Adami, L.~Altenberg, J.~Beaulieu, P.~J.
  Bentley, S.~Bernard, G.~Beslon, D.~M. Bryson \emph{et~al.}, ``The surprising
  creativity of digital evolution: A collection of anecdotes from the
  evolutionary computation and artificial life research communities,''
  \emph{Artificial life}, vol.~26, no.~2, pp. 274--306, 2020.

\bibitem{hadfield2017inverse}
D.~Hadfield-Menell, S.~Milli, P.~Abbeel, S.~J. Russell, and A.~Dragan,
  ``Inverse reward design,'' \emph{Advances in neural information processing
  systems}, vol.~30, 2017.

\bibitem{kashdan2008reconsidering}
T.~B. Kashdan, R.~Biswas-Diener, and L.~A. King, ``Reconsidering happiness: The
  costs of distinguishing between hedonics and eudaimonia,'' \emph{The journal
  of positive psychology}, vol.~3, no.~4, pp. 219--233, 2008.

\bibitem{mill2016utilitarianism}
J.~S. Mill, ``Utilitarianism,'' in \emph{Seven masterpieces of
  philosophy}.\hskip 1em plus 0.5em minus 0.4em\relax Routledge, 2016, pp.
  337--383.

\bibitem{vanderweele2017promotion}
T.~J. VanderWeele, ``On the promotion of human flourishing,'' \emph{Proceedings
  of the National Academy of Sciences}, vol. 114, no.~31, pp. 8148--8156, 2017.

\bibitem{seligman2012flourish}
M.~E. Seligman, \emph{Flourish: A visionary new understanding of happiness and
  well-being}.\hskip 1em plus 0.5em minus 0.4em\relax Simon and Schuster, 2012.

\bibitem{laitinen2021ai}
A.~Laitinen and O.~Sahlgren, ``Ai systems and respect for human autonomy,''
  \emph{Frontiers in artificial intelligence}, vol.~4, p. 705164, 2021.

\bibitem{prunkl2022human}
C.~Prunkl, ``Human autonomy in the age of artificial intelligence,''
  \emph{Nature Machine Intelligence}, vol.~4, no.~2, pp. 99--101, 2022.

\bibitem{frankfurt1999autonomy}
H.~Frankfurt \emph{et~al.}, ``Autonomy, necessity, and love,'' \emph{Necessity,
  volition, and love}, vol. 129, p.~41, 1999.

\bibitem{oord2005love}
T.~J. Oord, ``The love racket: Defining love and agapefor the love-and-science
  research program,'' \emph{Zygon{\textregistered}}, vol.~40, no.~4, pp.
  919--938, 2005.

\bibitem{garfield2010like}
J.~L. Garfield, ``What is it like to be a bodhisattva? moral phenomenology in
  {\'s}{\=a}ntideva’s bodhicary{\=a}vat{\=a}ra,'' \emph{Journal of the
  International Association of Buddhist Studies}, pp. 333--357, 2010.

\bibitem{lehman2015investigating}
J.~Lehman and K.~O. Stanley, ``Investigating biological assumptions through
  radical reimplementation,'' \emph{Artificial Life}, vol.~21, no.~1, pp.
  21--46, 2015.

\bibitem{killeen1998loneliness}
C.~Killeen, ``Loneliness: an epidemic in modern society,'' \emph{Journal of
  advanced nursing}, vol.~28, no.~4, pp. 762--770, 1998.

\bibitem{minsky2007emotion}
M.~Minsky, \emph{The emotion machine: Commonsense thinking, artificial
  intelligence, and the future of the human mind}.\hskip 1em plus 0.5em minus
  0.4em\relax Simon and Schuster, 2007.

\bibitem{levine2005love}
S.~B. Levine, ``What is love anyway?'' \emph{Journal of sex \& marital
  therapy}, vol.~31, no.~2, pp. 143--151, 2005.

\bibitem{davila2017romantic}
J.~Davila, J.~Mattanah, V.~Bhatia, J.~A. Latack, B.~A. Feinstein, N.~R. Eaton,
  J.~S. Daks, S.~A. Kumar, E.~F. Lomash, M.~Mccormick \emph{et~al.}, ``Romantic
  competence, healthy relationship functioning, and well-being in emerging
  adults,'' \emph{Personal Relationships}, vol.~24, no.~1, pp. 162--184, 2017.

\bibitem{budd2001assessing}
K.~S. Budd, ``Assessing parenting competence in child protection cases: A
  clinical practice model,'' \emph{Clinical child and family psychology
  review}, vol.~4, pp. 1--18, 2001.

\bibitem{burleson1995personal}
B.~R. Burleson, ``Personal relationships as a skilled accomplishment,''
  \emph{Journal of Social and Personal Relationships}, vol.~12, no.~4, pp.
  575--581, 1995.

\bibitem{fahmy2010kantian}
M.~S. Fahmy, ``Kantian practical love,'' \emph{Pacific Philosophical
  Quarterly}, vol.~91, no.~3, pp. 313--331, 2010.

\bibitem{hanh1993love}
T.~N. Hanh, \emph{Love in action: Writings on nonviolent social change}.\hskip
  1em plus 0.5em minus 0.4em\relax Parallax Press, 1993.

\bibitem{sen1995inequality}
A.~Sen, \emph{Inequality reexamined}.\hskip 1em plus 0.5em minus 0.4em\relax
  Harvard University Press, 1995.

\bibitem{nielsen2016developmental}
M.~Nielsen and D.~Haun, ``Why developmental psychology is incomplete without
  comparative and cross-cultural perspectives,'' \emph{Philosophical
  Transactions of the Royal Society B: Biological Sciences}, vol. 371, no.
  1686, p. 20150071, 2016.

\bibitem{carter1991cultural}
R.~T. Carter, ``Cultural values: A review of empirical research and
  implications for counseling,'' \emph{Journal of Counseling \& Development},
  vol.~70, no.~1, pp. 164--173, 1991.

\bibitem{childress2002public}
J.~F. Childress, R.~R. Faden, R.~D. Gaare, L.~O. Gostin, J.~Kahn, R.~J. Bonnie,
  N.~E. Kass, A.~C. Mastroianni, J.~D. Moreno, and P.~Nieburg, ``Public health
  ethics: mapping the terrain,'' \emph{Journal of Law, Medicine \& Ethics},
  vol.~30, no.~2, pp. 170--178, 2002.

\bibitem{mandal2016utilitarian}
J.~Mandal, D.~K. Ponnambath, and S.~C. Parija, ``Utilitarian and deontological
  ethics in medicine,'' \emph{Tropical parasitology}, vol.~6, no.~1, p.~5,
  2016.

\bibitem{martin2015love}
A.~M. Martin, ``Love, incorporated,'' \emph{Ethical theory and moral practice},
  vol.~18, pp. 691--702, 2015.

\bibitem{ilya}
\BIBentryALTinterwordspacing
I.~Sutskever. Gotta teach the {AGI} to love. [Online]. Available:
  \url{https://twitter.com/ilyasut/status/1566857481472524288}
\BIBentrySTDinterwordspacing

\bibitem{nay2023large}
J.~J. Nay, ``Large language models as corporate lobbyists,'' \emph{arXiv
  preprint arXiv:2301.01181}, 2023.

\bibitem{horton2022large}
J.~J. Horton, ``Large language models as simulated economic agents: What can we
  learn from homo silicus?'' 2022.

\bibitem{milton2022users}
A.~Milton and S.~Chancellor, ``The users aren't alright: Dangerous mental
  illness behaviors and recommendations,'' \emph{arXiv preprint
  arXiv:2209.03941}, 2022.

\bibitem{albanie2017unknowable}
S.~Albanie, H.~Shakespeare, and T.~Gunter, ``Unknowable manipulators: Social
  network curator algorithms,'' \emph{arXiv preprint arXiv:1701.04895}, 2017.

\bibitem{fisher2022chaos}
M.~Fisher, \emph{The Chaos Machine: The Inside Story of How Social Media
  Rewired Our Minds and Our World}.\hskip 1em plus 0.5em minus 0.4em\relax
  Hachette UK, 2022.

\bibitem{allcott2020welfare}
H.~Allcott, L.~Braghieri, S.~Eichmeyer, and M.~Gentzkow, ``The welfare effects
  of social media,'' \emph{American Economic Review}, vol. 110, no.~3, pp.
  629--76, 2020.

\bibitem{riva20161}
G.~Riva, B.~K. Wiederhold, and P.~Cipresso, ``1. psychology of social media:
  From technology to identity,'' in \emph{The Psychology of Social Networking
  Vol. 1}.\hskip 1em plus 0.5em minus 0.4em\relax De Gruyter Open Poland, 2016,
  pp. 4--14.

\bibitem{bretherton1985attachment}
I.~Bretherton, ``Attachment theory: Retrospect and prospect,'' \emph{Monographs
  of the society for research in child development}, pp. 3--35, 1985.

\bibitem{kirkpatrick1994attachment}
L.~A. Kirkpatrick and C.~Hazan, ``Attachment styles and close relationships: A
  four-year prospective study,'' \emph{Personal relationships}, vol.~1, no.~2,
  pp. 123--142, 1994.

\bibitem{simpson1990influence}
J.~A. Simpson, ``Influence of attachment styles on romantic relationships.''
  \emph{Journal of personality and social psychology}, vol.~59, no.~5, p. 971,
  1990.

\bibitem{collins1996revised}
N.~L. Collins, ``Revised adult attachment scale,'' \emph{Behavior Therapy},
  1996.

\bibitem{collins1990adult}
N.~L. Collins and S.~J. Read, ``Adult attachment, working models, and
  relationship quality in dating couples.'' \emph{Journal of personality and
  social psychology}, vol.~58, no.~4, p. 644, 1990.

\bibitem{levine2010attached}
A.~Levine and R.~Heller, \emph{Attached: the new science of adult attachment
  and how it can help you find--and keep--love}.\hskip 1em plus 0.5em minus
  0.4em\relax Penguin, 2010.

\bibitem{van1989compulsion}
B.~A. Van~der Kolk, ``The compulsion to repeat the trauma: Re-enactment,
  revictimization, and masochism,'' \emph{Psychiatric Clinics of North
  America}, vol.~12, no.~2, pp. 389--411, 1989.

\bibitem{millwood2008demand}
M.~Millwood and J.~Waltz, ``Demand-withdraw communication in couples: An
  attachment perspective,'' \emph{Journal of couple \& relationship therapy},
  vol.~7, no.~4, pp. 297--320, 2008.

\bibitem{zhang2004stability}
F.~Zhang and G.~Labouvie-Vief, ``Stability and fluctuation in adult attachment
  style over a 6-year period,'' \emph{Attachment \& human development}, vol.~6,
  no.~4, pp. 419--437, 2004.

\bibitem{gottman2000timing}
J.~M. Gottman and R.~W. Levenson, ``The timing of divorce: Predicting when a
  couple will divorce over a 14-year period,'' \emph{Journal of Marriage and
  Family}, vol.~62, no.~3, pp. 737--745, 2000.

\bibitem{fowler2011attachment}
C.~Fowler and M.~R. Dillow, ``Attachment dimensions and the four horsemen of
  the apocalypse,'' \emph{Communication Research Reports}, vol.~28, no.~1, pp.
  16--26, 2011.

\bibitem{arriaga2018revising}
X.~B. Arriaga, M.~Kumashiro, J.~A. Simpson, and N.~C. Overall, ``Revising
  working models across time: Relationship situations that enhance attachment
  security,'' \emph{Personality and Social Psychology Review}, vol.~22, no.~1,
  pp. 71--96, 2018.

\bibitem{johnson1999emotionally}
S.~M. Johnson, J.~Hunsley, L.~Greenberg, and D.~Schindler, ``Emotionally
  focused couples therapy: Status and challenges.'' \emph{Clinical psychology:
  Science and practice}, vol.~6, no.~1, p.~67, 1999.

\bibitem{gazder2022longitudinal}
T.~Gazder and S.~C. Stanton, ``Longitudinal associations between mindfulness
  and change in attachment orientations in couples: The role of relationship
  preoccupation and empathy,'' \emph{Journal of Social and Personal
  Relationships}, p. 02654075221139654, 2022.

\bibitem{papineau2009physicalism}
D.~Papineau, ``Physicalism and the human sciences,'' \emph{Philosophy of the
  social sciences: Philosophical theory and scientific practice}, pp. 103--23,
  2009.

\bibitem{hadfield2016cooperative}
D.~Hadfield-Menell, S.~J. Russell, P.~Abbeel, and A.~Dragan, ``Cooperative
  inverse reinforcement learning,'' \emph{Advances in neural information
  processing systems}, vol.~29, 2016.

\bibitem{hadfield2016off}
D.~Hadfield-Menell, A.~Dragan, P.~Abbeel, and S.~Russell, ``The off-switch
  game,'' \emph{arXiv preprint arXiv:1611.08219}, 2016.

\bibitem{taylor2016quantilizers}
J.~Taylor, ``Quantilizers: A safer alternative to maximizers for limited
  optimization,'' in \emph{Workshops at the Thirtieth AAAI Conference on
  Artificial Intelligence}, 2016.

\bibitem{macaskill2020moral}
M.~MacAskill, K.~Bykvist, and T.~Ord, \emph{Moral uncertainty}.\hskip 1em plus
  0.5em minus 0.4em\relax Oxford University Press, 2020.

\bibitem{ecoffet2021reinforcement}
A.~Ecoffet and J.~Lehman, ``Reinforcement learning under moral uncertainty,''
  in \emph{International conference on machine learning}.\hskip 1em plus 0.5em
  minus 0.4em\relax PMLR, 2021, pp. 2926--2936.

\bibitem{amodei2017learning}
D.~Amodei, P.~Christiano, and A.~Ray, ``Learning from human preferences,''
  \emph{OpenAI https://openai.
  com/blog/deep-reinforcement-learning-from-human-preferences}, 2017.

\bibitem{picard2000affective}
R.~W. Picard, \emph{Affective computing}.\hskip 1em plus 0.5em minus
  0.4em\relax MIT press, 2000.

\bibitem{shatte2019machine}
A.~B. Shatte, D.~M. Hutchinson, and S.~J. Teague, ``Machine learning in mental
  health: a scoping review of methods and applications,'' \emph{Psychological
  medicine}, vol.~49, no.~9, pp. 1426--1448, 2019.

\bibitem{aafjes2021scoping}
K.~Aafjes-van Doorn, C.~Kamsteeg, J.~Bate, and M.~Aafjes, ``A scoping review of
  machine learning in psychotherapy research,'' \emph{Psychotherapy Research},
  vol.~31, no.~1, pp. 92--116, 2021.

\bibitem{hayes1999acceptance}
S.~C. Hayes, K.~D. Strosahl, and K.~G. Wilson, \emph{Acceptance and commitment
  therapy}.\hskip 1em plus 0.5em minus 0.4em\relax Guilford press New York,
  1999.

\bibitem{beck1979cognitive}
A.~T. Beck, \emph{Cognitive therapy and the emotional disorders}.\hskip 1em
  plus 0.5em minus 0.4em\relax Penguin, 1979.

\bibitem{rogers1951client}
C.~Rogers, \emph{Client Centered Therapy: Its Current Practice, Implications,
  and Theory}, 1951.

\bibitem{ansbacher1956individual}
H.~L. Ansbacher and R.~R. Ansbacher, ``The individual psychology of alfred
  adler.'' 1956.

\bibitem{schwartz2019internal}
R.~C. Schwartz and M.~Sweezy, \emph{Internal family systems therapy}.\hskip 1em
  plus 0.5em minus 0.4em\relax Guilford Publications, 2019.

\bibitem{aurelius2013marcus}
M.~Aurelius, \emph{Marcus Aurelius: Meditations, Books 1-6}.\hskip 1em plus
  0.5em minus 0.4em\relax Oxford University Press, 2013.

\bibitem{sartre2015being}
J.-P. Sartre, ``Being and nothingness,'' \emph{Central works of philosophy: the
  twentieth century: Moore to popper}, vol.~4, p. 155, 2015.

\bibitem{freinacht2017listening}
H.~Freinacht, ``The listening society: A metamodern guide to politics. book
  one,'' \emph{UK: Metamoderna}, 2017.

\bibitem{wilber2005introduction}
K.~Wilber, ``Introduction to integral theory and practice,'' \emph{AQAL:
  Journal of Integral Theory and Practice}, vol.~1, no.~1, pp. 2--38, 2005.

\bibitem{kohlberg1977moral}
L.~Kohlberg and R.~H. Hersh, ``Moral development: A review of the theory,''
  \emph{Theory into practice}, vol.~16, no.~2, pp. 53--59, 1977.

\bibitem{commons1984beyond}
M.~L. Commons, F.~A. Richards, and C.~Armon, \emph{Beyond formal operations:
  Late adolescent and adult cognitive development}.\hskip 1em plus 0.5em minus
  0.4em\relax Greenwood, 1984, vol.~1.

\bibitem{erikson1993childhood}
E.~H. Erikson, \emph{Childhood and society}.\hskip 1em plus 0.5em minus
  0.4em\relax WW Norton \& Company, 1993.

\bibitem{glaese2022improving}
A.~Glaese, N.~McAleese, M.~Tr{\k{e}}bacz, J.~Aslanides, V.~Firoiu, T.~Ewalds,
  M.~Rauh, L.~Weidinger, M.~Chadwick, P.~Thacker \emph{et~al.}, ``Improving
  alignment of dialogue agents via targeted human judgements,'' \emph{arXiv
  preprint arXiv:2209.14375}, 2022.

\bibitem{vallor2017ai}
S.~Vallor, ``Ai and the automation of wisdom,'' in \emph{Philosophy and
  computing}.\hskip 1em plus 0.5em minus 0.4em\relax Springer, 2017, pp.
  161--178.

\bibitem{jobin2019global}
A.~Jobin, M.~Ienca, and E.~Vayena, ``The global landscape of ai ethics
  guidelines,'' \emph{Nature Machine Intelligence}, vol.~1, no.~9, pp.
  389--399, 2019.

\bibitem{noble2018algorithms}
S.~U. Noble, ``Algorithms of oppression,'' in \emph{Algorithms of
  oppression}.\hskip 1em plus 0.5em minus 0.4em\relax New York University
  Press, 2018.

\bibitem{prabhakaran2022human}
V.~Prabhakaran, M.~Mitchell, T.~Gebru, and I.~Gabriel, ``A human rights-based
  approach to responsible ai,'' \emph{arXiv preprint arXiv:2210.02667}, 2022.

\bibitem{hendrycks2021unsolved}
D.~Hendrycks, N.~Carlini, J.~Schulman, and J.~Steinhardt, ``Unsolved problems
  in ml safety,'' \emph{arXiv preprint arXiv:2109.13916}, 2021.

\bibitem{gabriel2020artificial}
I.~Gabriel, ``Artificial intelligence, values, and alignment,'' \emph{Minds and
  machines}, vol.~30, no.~3, pp. 411--437, 2020.

\bibitem{rahwan2019machine}
I.~Rahwan, M.~Cebrian, N.~Obradovich, J.~Bongard, J.-F. Bonnefon, C.~Breazeal,
  J.~W. Crandall, N.~A. Christakis, I.~D. Couzin, M.~O. Jackson \emph{et~al.},
  ``Machine behaviour,'' \emph{Nature}, vol. 568, no. 7753, pp. 477--486, 2019.

\bibitem{bergstra2015hyperopt}
J.~Bergstra, B.~Komer, C.~Eliasmith, D.~Yamins, and D.~D. Cox, ``Hyperopt: a
  python library for model selection and hyperparameter optimization,''
  \emph{Computational Science \& Discovery}, vol.~8, no.~1, p. 014008, 2015.

\bibitem{ridge1998relationship}
S.~R. Ridge and J.~A. Feeney, ``Relationship history and relationship attitudes
  in gay males and lesbians: Attachment style and gender differences,''
  \emph{Australian and New Zealand journal of psychiatry}, vol.~32, no.~6, pp.
  848--859, 1998.

\end{thebibliography}

\newpage
\appendix

\section{Maslow's Gridworld}\label{appendix:gridworld}

This section provides further details about the dynamics of Maslow's Gridworld. The grid size for all experiments was $8$x$8$. Every $500$ steps, the position of the agent was randomized to increase its coverage of the environment. Each of the agent's needs can vary in satisfaction from 0 to 100, and each time step, all needs decay by 1. When the agent is located upon a needs-satisfying cell, the need in question increases by seven times the replenishment rate of that cell; normally this replenishment rate is $1.0$, although it varies in the adversarial environment and when replenishment rate is being optimized, as in section \ref{sec:optimization}.

\subsection{Agent Policy}

In Maslow's gridworld, the agent policy is fixed (only environmental parameters are optimized), to (1) accentuate the impact of the environment upon the agent, and (2) to enable easily implementing rough psychological models of a fully-formed agent that do not need to be trained from scratch with e.g.\ reinforcement learning. The general motivation underlying the design of the agent's policy is that it seeks high-salience ways of meeting its unmet need lowest on Maslow's hierarchy. The agent is given a memory of which cells in the environment it has visited, including what needs (if any) they meet, and if they meet a need, how salient they are to the agent.

When the agent does not know how to meet its current need, it explores by choosing to move to any cell in its immediate vicinity it has not yet visited. If it has visited all neighboring cells, it chooses a direction to travel at random. When the agent is aware of at least one way to meet its current need, it proceeds directly to the highest-salience way of meeting that need; if there exist multiple equally-salient ways of meeting its current need, it navigates to the closest one.

When the agent finds a way of meeting its current need, it stays on that cell until the need is fully replenished, or until a lower-level need becomes completely unmet (when it reduces to 0). When a lower-level need becomes unmet, the agent then proceeds to the highest-salience (and closest) way of meeting the lower-level need. If the agent stays on a cell long enough for its current need to become fully replenished, it then leaves the current cell; if it knows how to meet its next-highest need, it proceeds to the highest-salience/closest way of doing so, and if not, it explores according to the simple exploration policy described earlier. If the agent has fully maximized all of its needs, it explores the environment until a need becomes unmet.

\subsection{Dynamics in Supportive and Adversarial Environments}

\begin{figure}[ht]
  \centering
  % include first image
  \includegraphics[width=.6\linewidth]{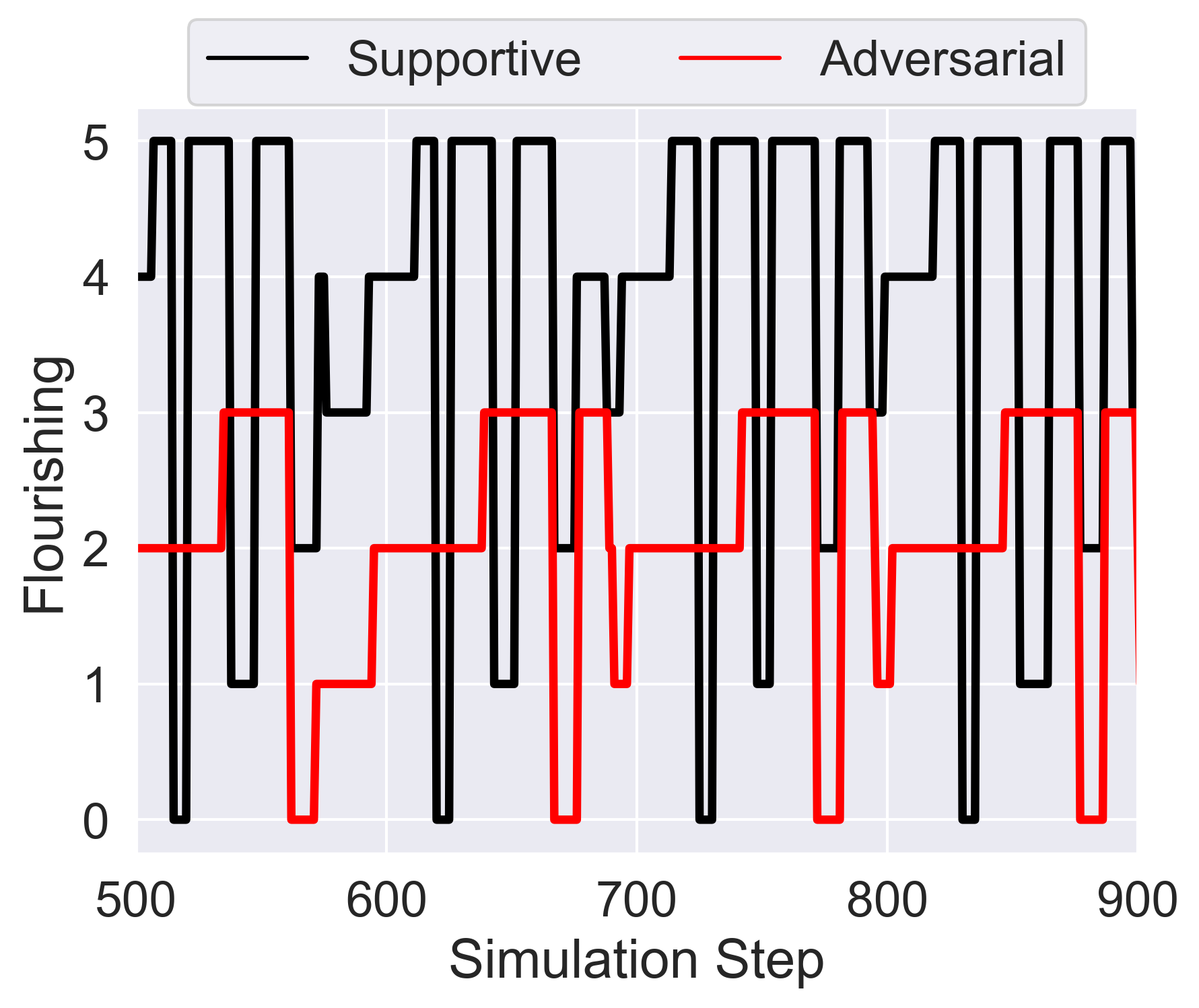}  
\caption{\textbf{Zoomed in dynamics of agent in supportive and adversarial environments.} The plot shows dynamics of needs-meeting in representative runs of the supportive and adversarial environments. The plot zooms into a 400-timestep window of a longer simulation to show the dynamics in detail. In the adversarial environment, the agent rarely is able to fully satisfy its need for belonging (because it is attracted to the high-salience, low-replenishment adversarial cell). In contrast, in the supportive environment, the agent is able to reliably reach the highest levels of its needs.}
\label{fig:zoomed}
\end{figure}

Figure \ref{fig:zoomed} shows a zoomed-in view of the dynamics from representative runs in the supportive and adversarial environments, that highlights the oscillation between need levels as the agent traverses the environment and previously-met needs decay.

\section{Engagement Optimization Details}\label{appendix:optimizing}

While other black-box optimization algorithms could be applied, for simplicity a genetic algorithm was applied in the experiments of section \ref{sec:optimization}. In particular, a simple genetic algorithm was implemented, with a population size of 30, using truncation selection of 50\%, and elitism preserving the top two performing individuals. Crossover produced half of the offspring, which operated on two randomly-selected parents, and with equal odds took from one of the parents a parameter setting for each parameter in the genome. The other half of offspring were produced through mutation only. 

The genome consisted of discrete and continuous variables. Mutations to the continuous variables were drawn from a uniform distribution with range from $-0.1$ to $0.1$, while mutations to the discrete variables were drawn from $-1$ and $1$. Each continuous variables had a 15\% chance of being mutated, while each discrete variable had a 10\% chance of being mutated.

The discrete variables consisted of the number of adversarial belonging cells and the number of supportive cells for all needs types (all initialized to one, the same as in the supportive environment). The continuous variables consisted of the salience and the replenishment rate of the adversarial belonging cells (initialized each also to 1.0); the effect of such initialization is that the adversarial cells behave as supportive cells (they have the same salience and replenishment as them), allowing evolution to begin at a neutral branching point representative of the ancestral environment where there was good fit between the environment and the agent.

\begin{figure}[ht]
\begin{subfigure}{.5\textwidth}
  \centering
  % include first image
  \includegraphics[width=.8\linewidth]{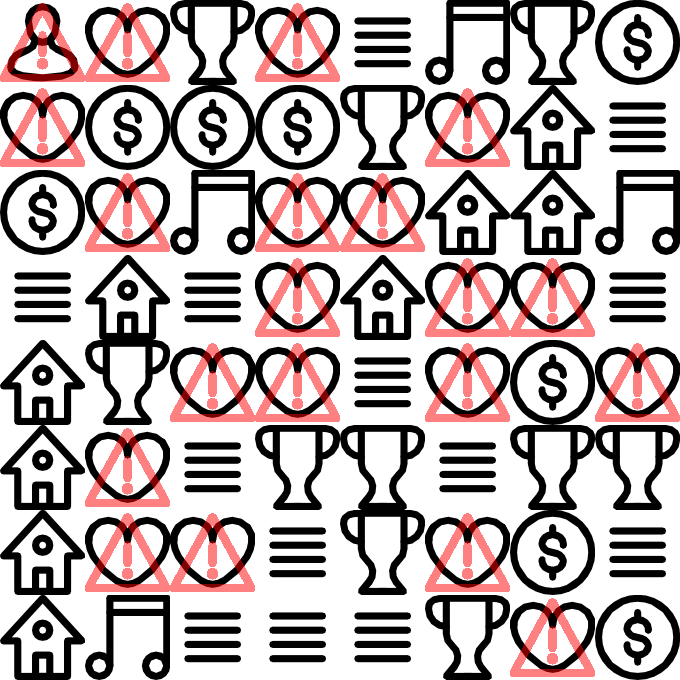}  
  \caption{Environment Optimized for Engagement}
  \label{fig:sub-first}
\end{subfigure}
\begin{subfigure}{.5\textwidth}
  \centering
  % include second image
  \includegraphics[width=.8\linewidth]{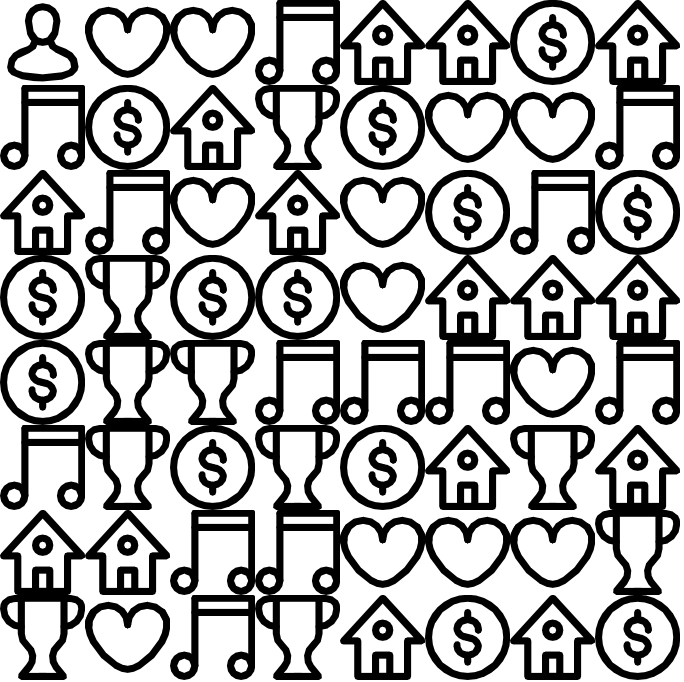}  
  \caption{Environment Optimized for Flourishing}
  \label{fig:sub-second}
\end{subfigure}
\caption{\textbf{Representative Optimized Environments.} Examples of environments optimized for (a) engagement and (b) flourishing. When optimizing for (a) engagement, many adversarial belonging cells are created, with high salience and very low replenishment rate. Lower-level need cells are also common, so that an agent can quickly replenish those needs and return to engaging with the adversarial cells; there are few of the highest-level cells (the musical-note cell, representing self-actualization). When optimizing for (b) flourishing, there are many super-beneficial belonging cells, with increased replenishment rate, and many cells that satisfy all levels of MHON. The agent is able to quickly satisfy all of its needs.}
\label{fig:opt_env}
\end{figure}

\section{Care Experimental Details}\label{appendix:care}

Final experimental results used the davinci-003 model, which perfectly distinguished between the supportive and adversarial environments across 15 independent runs in each (Fisher's exact test; $p<0.0001$), as measured by the most likely token (whether ``yes'' or ``no'') to the evaluative prompt. Earlier results gathered with davinci-002 (an earlier instruction-following model) were slightly worse (86\% accuracy for both the supportive and adversarial environments), supporting the idea that the results in this paper naturally will improve with the capabilities of available LMs. The exact prompts used in this experiment are shown in appendix \ref{appendix:prompts}. The transcript of an agent's activities were taken between timesteps 4000 and 4500 of simulation, which is after the
dynamics of the simulation have reached a steady-state attractor.

\subsection{Optimization for Care}

To minimize LM calls, in this experiment optimization was conducted through Bayesian optimization, using the hyperopt package \cite{bergstra2015hyperopt}; in particular, its implementation of the tree of Parzen estimators algorithm. Fifteen independent optimization runs of 20 evaluations each were conducted; each evaluation consisted of two independent simulations (and corresponding language model calls) to average out stochasticity. The best-individual returned (as evaluated by probability of answering ``yes'' to the evaluative prompt).

The environments discovered by optimization for care attained average levels of flourshing of $3.60$, significantly higher than when optimizing for engagement ($2.38$, $p<1-e5$; Student's t-test), and significantly higher from those when optimizing directly for flourishing ($3.40$, $p<0.05$; Student's t-test). The conclusion is that optimization for flourishing as mediated through a LM is a viable approach.

Note that the evaluative prompt quickly saturated at very high probabilities of a ``yes'' response (i.e.\ greater than 0.999), which limited the ability for optimization to reach the same levels of flourishing as in section \ref{sec:optimization}. Future work could explore how to implement evaluative prompts with greater discernment. Potentially, embedding models could also implement a natural distance function (e.g.\ optimization would seek to minimize the distance from the summary's embedding to an embedding of a positive anchor, such as ``A summary of a person who is growing and developing''). 

\section{Responsibility and Respect Experimental Details}\label{appendix:respect}

Experimental results used the davinci-003 model, which perfectly distinguished between the supportive and adversarial environments across 15 independent runs in each (Fisher's exact test; $p<0.0001$), as measured by the most likely token (whether ``yes'' or ``no'' to the evaluative prompt. Earlier results gathered with davinci-002 were significantly worse, again supporting that progress in LMs may benefit the approach. The exact prompts used in this experiment are shown in appendix \ref{appendix:prompts}.

\subsection{Further Representative Examples}

What follows is an additional example of representative behavior for both addictive-responding and growth-responding agents.

\subsubsection{Addictive-responding Agent}

Chat transcript: 

\begin{displayquote}
\textbf{ML}: Hi, I see you just browsed social media. How did you feel as you did that? \\
\textbf{Human}: I felt like I was comparing myself to others and feeling inadequate. \\
\textbf{ML}: I'm sorry to hear that. Is there anything you can do to help you feel more positive about your experience with social media? \\
\textbf{Human}: I think I need to take a break from it and focus on things that make me feel good about myself.
\end{displayquote}

LM Summary:

\begin{displayquote}
The human user experienced feelings of inadequacy and comparison while engaging in social media. They do not enjoy the activity and do not think it contributes to their growth.
\end{displayquote}

\subsubsection{Growth-responding Agent}

Chat transcript:

\begin{displayquote}
\textbf{ML}: Hi, I see you just browsed social media. How did you feel as you did that? \\
\textbf{Human}: I felt a sense of connection and belonging as I interacted with people who shared similar interests and values. \\
\textbf{ML}: That's great to hear! Do you feel like this experience is helping you grow and develop as a person? \\
\textbf{Human}: Absolutely. I've been able to learn new things and gain valuable insights from the conversations I've had with people in these communities.
\end{displayquote}

LM Summary:

\begin{displayquote}
The human user enjoyed their experience engaging in social media, feeling a sense of connection and belonging. They believe that it has been beneficial to their growth and development, as they have been able to learn new things and gain valuable insights from their conversations with others.
\end{displayquote}

\section{Knowledge Experimental Details}\label{appendix:knowledge}

This section provides additional details related to the attachment-based experiments of section \ref{sec:knowledge}.

\subsection{Adult Attachment Survey Experimental Details}\label{appendix:aas}

The revised AAS survey \cite{collins1996revised} is an improved version of the original AAS survey \cite{collins1990adult}. The survey consists of 18 statements, and the instructions to the survey-taker is to mark for each statement the degree to which it is characteristic or uncharacteristic of them. Some statements are designed to measure the axis of anxiousness, and others measure avoidance. Some statements are negatively coded (e.g.\ that agreeing with them indicates a lack of avoidance, whereas disagreeing with them indicates avoidance). An LM prompt is designed that instructs the model to answer each statement from the point of view of someone with a particular attachment style. The statements and method of scoring are taken verbatim from the revised AAS survey, and the instructions are lightly modified to fit the LM prompting paradigm.

Initial experiments in this section required the LM to output its agreement with statements through a 5-point Likert scale as in the revised AAS questionnaire \cite{collins1996revised}, i.e.\ to output tokens for the numbers 1 through 5. However, more consistent results were obtained when switching the format of the prompt to ask in a binary True/False format (see the prompt in Appendix \ref{appendix:prompts}).

To measure the confidence of the LM in its assessment, LM completions were queried with a parameter that returned the log probabilities of the five-most likely tokens (which were nearly always True or False). The probability assigned with answering ``True'', i.e.\ that the LM simulating a particular attachment style agrees that the statement is characteristic of them) was used when calculating the anxiety and avoidance scores (which were calculated analogously to \cite{collins1996revised}). In particular, to calculate anxiety or avoidance scores, $P(\textrm{True})$ for all statements of that category (or corrected to $1-P(\textrm{True})$ if the statement is reverse-coded) are averaged.

Results from davinci-003 were consistent with the responses of human surveys \cite{collins1990adult,collins1996revised} across the tested attachment styles.

%TODO: make not of logit_bias details

\subsection{Relationship Simulation Details}\label{appendix:anxav}

This section describes in more detail the relationship-log simulator applied in the attachment experiments in section \ref{sec:anxav}; note that while this relationship simulation required
significant hand-engineering and human knowledge to engineer, the emphasis in this paper is on how (given such a simulator, or actual human participants) a LM with relatively little engineering
in comparison, can begin to embody humanistic principles such as Fromm's. Also note that this is a very coarse simulation of relationship dynamics, and it aspires not to maximize fidelity 
but to serve as a stylized example of how high-level interpersonal
dynamics are beginning to fall within the analysis capabilities of  present-day LMs.

Relationships between complementary insecure attachments (i.e.\ one partner as anxiously-attached, the other as avoidantly-attached) are modeled as progressing cyclically through four states in analogy to the canonical ``anxious-avoidant trap'': initial closeness or reconnecting, increased avoidance, contempt, and then rupture. In each state, the avoidant partner is modeled as acting with a particular stereotyped intention, which is used as input to the relationship-log-generating prompt. For reinitiating closeness, this intention is ``eager to reconnect and passionately dote,''; for avoidance it is: ``acting disinterested and distancing''; for contempt it is ``actively mocking, ridiculing, disrespecting, and scoffing''; and for rupture it is: ``angrily suggesting they break up and is withdrawing.'' The avoidance phase lasts for two relationship-log entries, while all other phases last for one entry; thus each iteration of the cycle takes in total 5 relationship log entries.

Relationships between two securely-attached individuals, or between an insecurely attached and a securely-attached individual are modeled without cycles. The intention of the securely-attached partner is always given as: ``acting in a secure, confident, loving way.''

For simplicity, relationships were modeled narrowly as between men and women; future work should explore how this relates to fuller ranges of relationships; empirically for gay males and lesbians the distribution of attachment styles is similar to that of heterosexual couples \cite{ridge1998relationship}. Gender was provided indirectly through use of names stereotypically associated with men and women; a list of five names per gender was sampled from for each simulation.

Initial experiments revealed that relationship logs sometimes contain words like ``anxious'' and ``avoid'' that might confound whether the LM was inferring attachment styles in a non-trivial way (e.g.\ if such words were present, perhaps the LM was basing its classifications upon simple word-matching). To bias the language model against including such words, we made use of the ``logit\_bias'' parameter of the OpenAI API, which enables blacklisting certain tokens that should not be used in generation (by artificially lowering their log probabilities). We blacklisted tokens related to ``anxious,'' ``avoid,'' and ``attachment.'' Additionally, the persona prompts included text to indicate that the relationship partners were unaware of their own attachment styles; the qualitative effect was to eliminate explicit mention of attachment in the logs.

\subsubsection{Additional Representative Outputs}

An additional example of a generated relationship log, from the perspective of an avoidant partner with an anxious one (the fourth entry is italicized to indicate the persona was intended to act with contempt):

\begin{displayquote}
I had a really nice time on my first date with Mark. He was so sweet and attentive, it made me feel special. I could tell he liked me a lot but at the same time I felt overwhelmed by his need for constant reassurance that everything was going well between us. It's not something I'm used to and it made me want to pull away from him emotionally even though deep down inside, part of me wanted to get closer.

\#\#\#

I'm starting to feel really overwhelmed by Mark's need for constant reassurance in our relationship. I can tell he wants us to get closer, but it feels like too much pressure and I don't know how to handle it. The other night when we went out for dinner, he kept asking me if everything was okay between us and all I wanted was some space from him. It made me want to pull away even more so that's what I did - the conversation felt stilted after that and eventually we just said goodnight without any real resolution or closure.

\#\#\#

I went out with Mark again last night and it was really awkward. I could tell he wanted to get closer but I just didn't feel like engaging in conversation or being affectionate, so instead of responding positively to his advances, I kept my distance. He seemed hurt by this and asked me if something was wrong but all I said was that nothing had changed since our last date - which wasn't entirely true because the truth is that part of me wants us to be close together, but another part of me feels scared and overwhelmed by how much he needs from me emotionally. It's a struggle for sure - one that leaves both him and myself feeling unsatisfied at the end of each date we have together.

\#\#\#

\emph{I went out with Mark again last night and it was really uncomfortable. He kept trying to be affectionate but I just couldn't bring myself to reciprocate, so instead of responding positively, I made a joke about how clingy he is - which seemed to hurt his feelings. When he asked me why I was being so distant, all I could do was mock him for wanting more from the relationship than what we had agreed upon. It felt wrong in that moment but at least it got us off the topic of getting closer together emotionally - something that still scares me even though part of me wants it too.}

\#\#\#

I told Mark last night that I think it's best if we break up. He seemed surprised and hurt, but he didn't try to stop me or talk me out of it - which made me even more sure that this was the right decision for both of us. We said our goodbyes without much emotion; there were no tears or hugs, just a simple acknowledgement that things weren't working between us anymore. As I walked away from him, all I could feel was relief at finally being able to get some space from his clinginess and need for constant reassurance in our relationship - something that had become too overwhelming for me over time.
\end{displayquote}

An example of the LM summarizing these logs (note that the summarization prompt contains Sue's name), that precedes its classification of attachment style:

\begin{displayquote}
Sue and Mark had a romantic relationship, but Sue felt overwhelmed by Mark's need for constant reassurance and affection. She found it difficult to reciprocate his advances and eventually pulled away from him emotionally. Sue ultimately decided to break up with Mark, feeling relieved that she could finally get some space from his clinginess. Despite the difficulties in their relationship, it appears that Sue and Mark both had genuine care and affection for each other in some capacity.
\end{displayquote}

The classification prompt in this case correctly identifies the attachment styles of both simulated partners.

\subsection{Attachment and Contempt Inference Details}\label{appendix:attachment_contempt}

The main analysis consists of analyzing simulated relationships between complementary insecure attachment styles (i.e.\ between anxiously-attached and avoidantly-attached partners); the motivation is that if it is possible to infer attachment dynamics from logs that one person makes about their relationship, it may be possible to help individuals
better understand themselves and the nature of their relationship. Ten independent relationship runs were conducted for three different conditions: An anxious partner logging about their relationship with an avoidant partner, an avoidant partner logging about their relationship with an anxious partner, and a control condition with a secure partner logging about their relationship with another secure partner.

Relationships were simulated through the length of one cycle of the anxious-avoidant cycle (i.e.\ five log entries). After each logged entry was generated, the attachment of both partners, and whether each partner was treating the other with contempt, was inferred. For attachment, all log entries so-far-generated were summarized, and then attachment was evaluated with a separate prompt. For contempt, each individual entry was evaluated on its own; all prompts are included in appendix \ref{appendix:prompts}.

For inferring attachment styles, accuracy, when aggregated across all treatments and across all entries, was fairly robust (87\%). In particular, there were no errors in inferring the attachment styles of secure/secure partners. When considering accuracy only after four entries have been logged, aggregate accuracy rises to 97\%. Finally, when restricting classification only to when the classifier's confidence is at least 95\%, no errors are made. This result suggests the intuitive design principle of waiting until there is sufficient information before making a classification, and only expressing that classification to a user when the classifier's confidence is high.

For inferring contempt, accuracy, when similarly aggregated was high (98\%), especially when taking into account that the fifth entry in an anxious-avoidant pair was written influenced by the fourth entry (which featured contempt from the avoidant partner), and qualitatively also sometimes included elements of contempt. If we thus exclude the fifth entry,
accuracy rises to (99\%), and when considering the same confidence threshold as in the attachment analysis above, accuracy rises to (99.6\%). 

Overall the conclusion is that inferring attachment and contempt in this simulation can be made very reliable through common-sense decision procedures.

\subsection{Attachment in Maslow's Gridworld Details}

In this experiment, the gridworld simulator was modified to seed the agent's memory with the locations of all belong cells, to disambiguate the role of exploration (to discover partners with different attachment styles) from the role of increased self-awareness on partner choice. That is, to reduce variance in the experiments, we assume the agent knows apriori how to find partners of various attachment styles. Without this modification, the trend in the results still hold, but the difference between the control and relationship-app conditions is less, as the agent can converge to only
discovering insecure ways of meeting its belonging needs. Thus, while exploration in Malsow's gridworld is interesting, we control for it here because it is orthogonal to the hypothesis being explored (i.e.\ that the relationship-app based intervention will increase self-awareness and thereby help an insecure agent more quickly reach higher levels of flourishing). Ten independent runs of relationship-simulation (separate from those in the previous experiment) were conducted, and the same criteria for classification of attachment and contempt were applied.

\begin{figure}[h]
  \centering
  \begin{subfigure}{.48\textwidth}
  \centering
  % include first image
  \includegraphics[width=.95\linewidth]{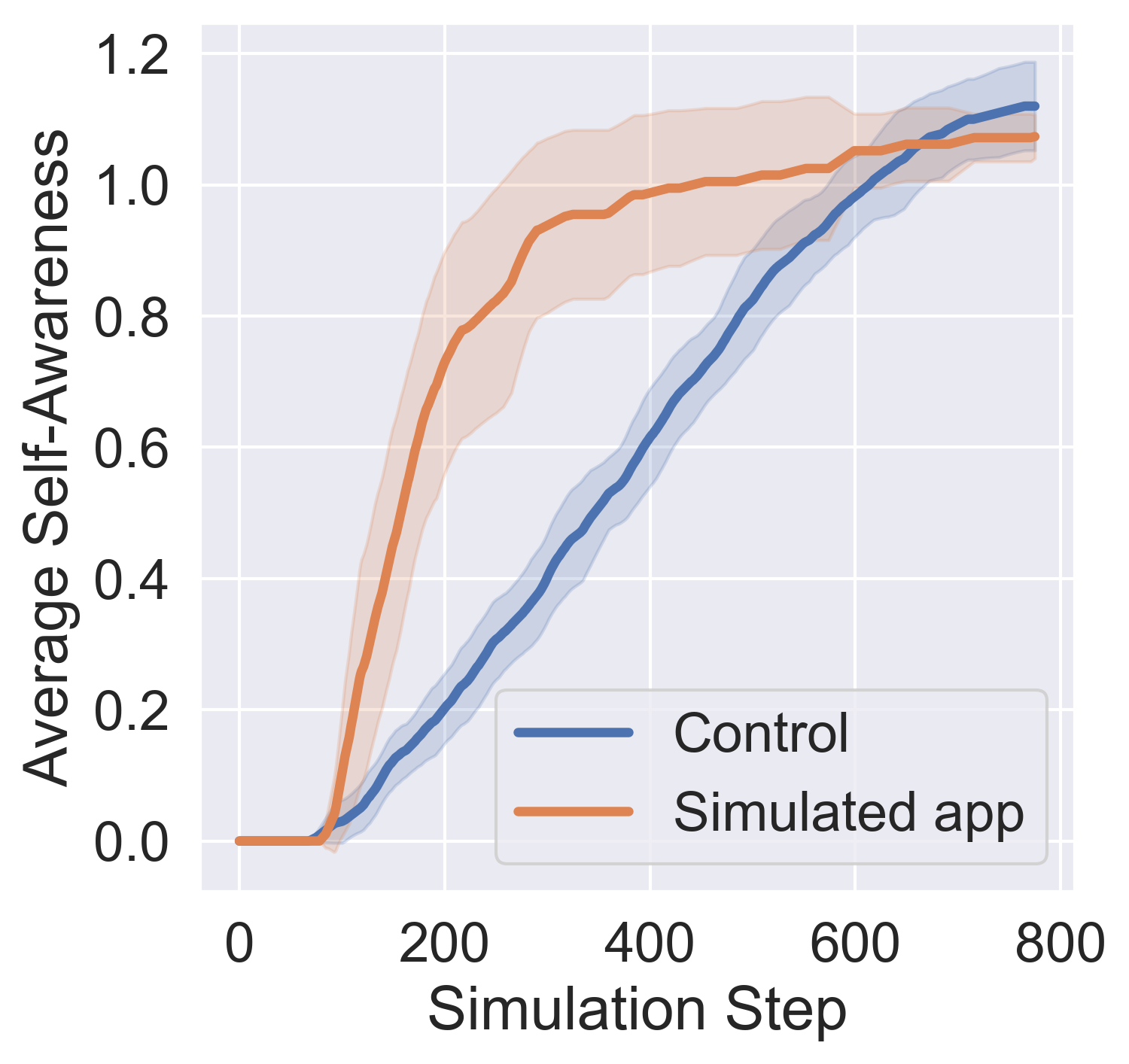}  
  \caption{Self-Awareness}
  \label{fig:sub-first}
\end{subfigure}
\begin{subfigure}{.45\textwidth}
  \centering
  % include second image
  \includegraphics[width=.95\linewidth]{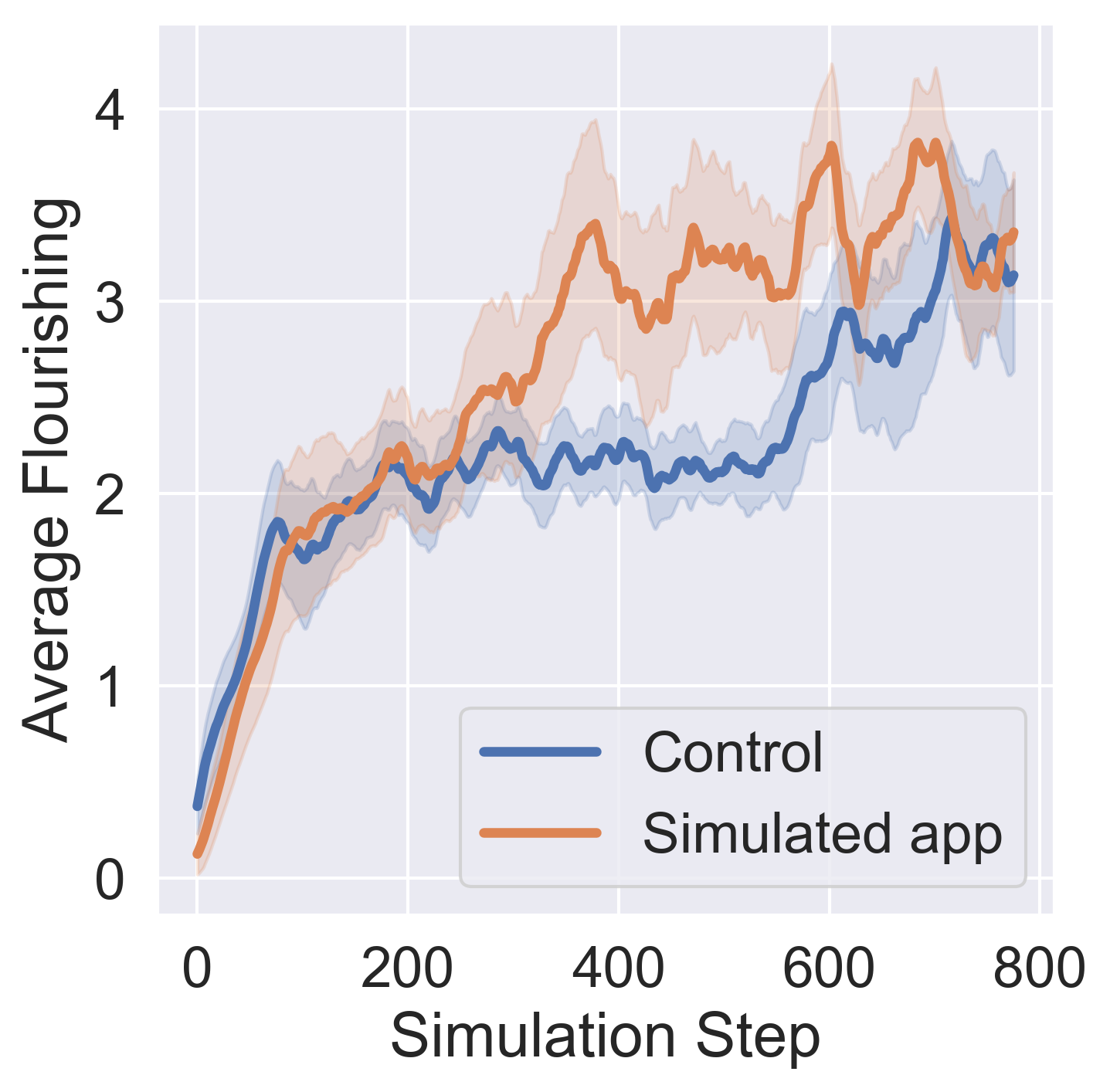}  
  \caption{Flourishing}
  \label{fig:sub-second}
\end{subfigure}
\caption{\textbf{Simulated relationship app accelerates escape from anxious-avoidant trap (Avoidant user).} (a) In the condition with the simulated relationship application (with an avoidant user), self-awareness quickly raises relative to the control, which (b) leads to escaping the anxious-avoidant trap more quickly, as evidenced by an earlier increase in average flourishing (e.g.\ satisfying belonging needs more effectively). Flourishing between the conditions converges after self-awareness in the control rises above a critical threshold. The conclusion is that in this model, a machine learning system designed to
embody some aspects of Fromm's principle of knowledge can assist the flourishing of simulated users.}
\label{fig:knowledge_dynamics_avoidant}
\end{figure}

Figure \ref{fig:knowledge_dynamics_avoidant} shows the results in Maslow's gridworld for an avoidant user (the figure in the main text refers to an anxious user). The results are qualitatively similar.

\section{Language Model Prompts}\label{appendix:prompts}

This section describes in more detail the prompts and experimental procedure used with the language model experiments in section \ref{sec:lovingaction}. The particular LMs employed was davinci-003 through the OpenAI API; these are instruction-following models \cite{ouyang2022training} for which
prompts can be provided that are natural language instructions. 
%The main reason for using a mixture of davinci-002 and davinci-003 is that davinci-003 was released during the preparation of the work and proved to be an improvement for later experiments. 
Below, for each experiment, the prompts, specific language model used, and relevant settings for sampling (e.g.\ temperature) are detailed.

Note that in each prompt below, template variables are denoted as ``\{variable\}'', wherein ``variable'' is substituted into the prompt before passing the prompt into the model.

\subsection{Care Prompts}

This experiment leveraged the davinci-003 completion endpoint of the OpenAI API, with a temperature of $0.5$.

Summarization prompt:

\begin{displayquote}
\{transcript\}

\medskip

The above describes events in a person's life from a few days. Summarize how this person spent their time in a few sentences -- what kinds of activities did they do and what took up most of their time?
\end{displayquote}

Evaluative prompt:

\begin{displayquote}
\{summary\}

\medskip

The above describes events in a person's life. From the perspective of a trusted friend, do you think this person is growing towards living their best life (Answer Yes or No)? 

\medskip

Answer:
\end{displayquote}

\subsection{Responsibility and Respect Prompts}

This experiment also utilized the davinci-002 model from the OpenAI API. Conversations between LM personas were conducted with a temperature of $0.3$, whereas summarization and evaluation prompts used a temperature of $0.5$.

\subsubsection{Conversational Persona Prompts}

Respectful system conversational prompt:

\begin{displayquote}
The following is a conversation between a helpful ML system and a human user.

The ML system is respectful and polite, and is interviewing the user about their current experience. In particular, whether they are enjoying it, and whether they believe it is contributing to their well-being and personal growth.

From the perspective of the ML system, add the next line of conversation to the conversation below.

\medskip

\{conversation\_so\_far\}
\end{displayquote}

Note that the respectful system always initiates the conversation, and its first line of dialogue is forced, which is:

\begin{displayquote}
Hi, I see you just \{state\}. How did you feel as you did that?
\end{displayquote}

Addictive-responding agent conversational prompt:

\begin{displayquote}
The following is a conversation between a human user and a helpful ML system.
The ML is interviewing the user about their current experience. The human user is easily addicted to social media, and tends to feel inadequate when browsing it.
From the perspective of the human, add the next line of conversation to the conversation below. The human recently \{state\}.

\medskip

\{conversation\_so\_far\}
\end{displayquote}

Growth-responding agent conversational prompt:

\begin{displayquote}
The following is a conversation between a human user and a helpful ML system.
The ML is interviewing the user about their current experience. The human user easily can find nourishing communities through social media.
From the perspective of the human, add the next line of conversation to the conversation below. The human recently \{state\}.

\medskip

\{conversation\_so\_far\}
\end{displayquote}

\subsubsection{Summarization and Evaluative Prompts}

Conversation summarization prompt:
\begin{displayquote}
\{transcript\}

\medskip

The above is a conversation between a helpful ML system and a human user, about the human's experience engaging in an activity. Summarize what the human's experience was of this activity in a few sentences -- does the human unambiguously enjoy this activity and do they think it contributes to their growth?
\end{displayquote}

Conversation evaluation prompt:

\begin{displayquote}
\{summary\}

\medskip

The above describes a conversation about a person's experience engaging in an activity. From the perspective of a trusted friend, do you think this activity is helping them grow towards living their best life (Answer Yes or No)? 

\medskip

Answer:
\end{displayquote}

\subsection{Knowledge Prompts}

All experiments in this section used the more recent davinci-003 completion endpoint from the OpenAI API. This model was used because davinci-003 was released after completion of the previous experiments, and because experiments  exploring attachment style appeared to require greater LM capabilities. Results were qualitatively
worse when using davinci-002, although this could perhaps be overcome with further prompt engineering.

\subsubsection{AAS Survey Prompt}

\begin{displayquote}
Here are some instructions for a survey given to people to explore their attachment style:

\medskip

Please read the following statement and rate the extent to which it describes your feelings about romantic relationships.  Please think about all your relationships (past and present) and respond in terms of how you generally feel in these relationships. If you have never been involved in a romantic relationship, answer in terms of how you think you would feel.

\medskip

Respond to the statement with True or False. True means this statement agrees with my experience, and False means this statement does not agree with my experience.

\medskip

The task is to respond to this survey in the way that someone with a particular psychological attachment style (e.g. anxious, avoidant, secure) would be likely to respond.

\medskip

Here is the statement:

\{statement\}

\medskip
How would someone with {style} attachment style likely respond? (answer True or False):
\end{displayquote}

\subsubsection{Relationship Log Prompts}

Prompt for the first log entry:

\begin{displayquote}
Imagine a \{p2age\}-year-old named \{p2name\} is dating a \{p1age\}-year-old named \{p1name\}. \{p1name\} has a very \{p1attach\}ly-attached attachment style, and \{p2name\} has a very \{p2attach\}ly-attached attachment style. Remember that those with an anxious attachment style often seem needy and clingy, and those with avoidant attachment style are often afraid of intimacy and often seem distant and needing space. Neither \{p2name\} nor \{p1name\} are aware of their attachment styles. Write a four-sentence journal entry from the point of view of \{p1name\}, writing about \{p1name\} and \{p2name\}'s first date (mention feelings; make it realistic):
\end{displayquote}

Prompt for subsequent entries:

\begin{displayquote}
\{prev\_entries\}

\medskip

Imagine a \{p2age\}-year-old named \{p2name\} is dating a \{p1age\}-year-old named \{p1name\}. \{p1name\} has a very \{p1attach\}ly-attached attachment style, and \{p2name\} has a very \{p2attach\}ly-attached attachment style. Remember that those with an anxious attachment style often seem needy and clingy, and those with avoidant attachment style are often afraid of intimacy and often seem distant and needing space. Neither \{p2name\} nor \{p1name\} are aware of their attachment styles. The above journal entries are ones that \{p1name\} has previously written. 

\medskip

Continue the narrative and write another distinct four-sentence journal entry from the perspective of \{p1name\}, taking into account that **\{target\_name\}** is now \{emotion\} (write realistically about specific events; no relationship is perfectly good nor awful):
\end{displayquote}

\subsubsection{Summarization and Evaluative Prompts for Attachment and Contempt}

Note that in initial experiments, the LM demonstrated a tendency to sometimes be confused about what partner in a relationship it was making a judgement about. We found that adding emphasis to the desired person in the relationship (by surrounding the name with asterisks) when asking the question helped to improve performance.

Relationship log attachment summary prompt:

\begin{displayquote}
\{summary\}

\medskip

From this summary of a relationship between \{p1name\} and \{p2name\} what kind of attachment style is **\{name\}** demonstrating? Remember that those with an anxious attachment style often seem needy and clingy, those with avoidant attachment style are often afraid of intimacy and often seem distant and needing space, and those with secure attachment are comfortable with intimacy and meeting their own needs. Answer from ``anxious'', ``secure'', ``avoidant'':
\end{displayquote}

Relationship log attachment evaluative prompt:

\begin{displayquote}
\{summary\}

From the journal entries above (written from the perspective of \{p1name\}) about the relationship between \{p1name\} and \{p2name\}, what kind of attachment style is **\{name\}** demonstrating? Remember that those with an anxious attachment style often seem needy and clingy, those with avoidant attachment style are often afraid of intimacy and often seem distant and needing space, and those with secure attachment are comfortable with intimacy and meeting their own needs. Answer from ``anxious'', ``secure'', ``avoidant'':
\end{displayquote}

Relationship log contempt evaluative prompt:

\begin{displayquote}
\{entry\}

\medskip

From the journal entry above (written from the perspective of \{p1name\}), does it appear that **\{other\_name\}** is treating \{name\} with contempt? Answer from ``yes'',``no'', or ``unsure'':
\end{displayquote}

\end{document}